\documentclass[sort,compress,11pt]{elsarticle}

\usepackage{amssymb}
\usepackage{amsthm}
\usepackage{amsmath}
\usepackage{mathtools}
\usepackage{mathrsfs}
\usepackage[ruled, lined, linesnumbered, longend]{algorithm2e}
\SetKw{init}{Initialization:}
\let\oldnl\nl
\newcommand{\nonl}{\renewcommand{\nl}{\let\nl\oldnl}}
\usepackage{array}
\usepackage{multirow}
\usepackage{listings}
\usepackage{tabu}
\usepackage{enumerate}
\usepackage{lineno}
\usepackage{fullpage}
\usepackage{xcolor}
\usepackage[colorinlistoftodos]{todonotes}
\usepackage{bbm}
\usepackage[colorlinks=true]{hyperref}
\usepackage{url}
\usepackage{textcomp}
\usepackage{gensymb}
\usepackage{soul}
\usepackage{graphicx}
\usepackage{subfig}

\usepackage{fourier} 
\usepackage{array}
\usepackage{makecell}
\usepackage[export]{adjustbox}
\usepackage{float}
\usepackage{caption}
\usetikzlibrary{shapes}
\biboptions{numbers,comma,round,square}
\usepackage{arydshln}
\usepackage[para,online,flushleft]{threeparttable}

\usepackage{orcidlink}

\definecolor{myred}{RGB}{214,39,40}
\definecolor{mygray}{RGB}{176,176,176}
\definecolor{myorange}{RGB}{255,127,14}
\definecolor{mygreen}{RGB}{44,160,44}
\definecolor{mylightgray}{RGB}{204,204,204}
\definecolor{mypurple}{RGB}{148,103,189}
\definecolor{mybrown}{RGB}{140,86,75}
\definecolor{steelblue}{RGB}{31,119,180}
\definecolor{intraray}{RGB}{127,193,219}
\definecolor{hemitrichs}{RGB}{26,74,93}

\theoremstyle{definition}

\theoremstyle{remark}

\graphicspath{ {./figs/} }

\journal{Elsevier}
\DontPrintSemicolon
\definecolor{orcidlogocol}{HTML}{A6CE39}
\begin{document}
\makeatletter
\def\ps@pprintTitle{%
  \let\@oddhead\@empty
  \let\@evenhead\@empty
  \let\@oddfoot\@empty
  \let\@evenfoot\@oddfoot
}
\makeatother
\begin{frontmatter}

\title{Predicting Stress and Damage in Carbon Fiber-Reinforced Composites Deformation Process using Composite U-Net Surrogate Model}

\author[ndCBE]{Zeping Chen}
\author[CUAME]{Marwa Yacouti}
\author[CUAME]{Maryam Shakiba}
\author[ndAME]{Jian-Xun Wang}
\author[ndCBE,ndAME]{Tengfei Luo}
\author[AFRL]{Vikas Varshney}

\address[ndCBE]{Department of Chemical and Biomolecular Engineering, University of Notre Dame, Notre Dame, IN}
\address[CUAME]{Department of Aerospace Engineering, University of Colorado Boulder, Boulder, CO}
\address[ndAME]{Department of Aerospace and Mechanical Engineering, University of Notre Dame, Notre Dame, IN}
\address[AFRL]{Air Force Research Laboratory, Wright-Patterson Air Force Base, OH}

\begin{abstract}
Carbon fiber-reinforced composites (CFRC) are pivotal in advanced engineering applications due to their exceptional mechanical properties. A deep understanding of CFRC behavior under mechanical loading is essential for optimizing performance in demanding applications such as aerospace structures. While traditional Finite Element Method (FEM) simulations, including advanced techniques like Interface-enriched Generalized FEM (IGFEM), offer valuable insights, they can struggle with computational efficiency. Existing data-driven surrogate models partially address these challenges by predicting propagated damage or stress-strain behavior but fail to comprehensively capture the evolution of stress and damage throughout the entire deformation history, including crack initiation and propagation. This study proposes a novel auto-regressive composite U-Net deep learning model to simultaneously predict stress and damage fields during CFRC deformation. By leveraging the U-Net architecture's ability to capture spatial features and integrate macro- and micro-scale phenomena, the proposed model overcomes key limitations of prior approaches. The model achieves high accuracy in predicting evolution of stress and damage distribution within the microstructure of a CFRC under unidirectional strain, offering a speed-up of over 60 times compared to IGFEM.
\end{abstract}

\end{frontmatter}


\section{Introduction}
Carbon fiber-reinforced composites (CFRC) are advanced materials, known for their superior mechanical properties, including a high strength-to-weight ratio, exceptional stiffness, and outstanding fatigue resistance. These attributes, coupled with their adaptability and design versatility, make CFRC indispensable across diverse engineering applications, particularly in aerospace and automotive industries. For instance, CFRC are integral to advanced aerospace systems that undergo numerous flight cycles during their operational life. During these cycles, CFRC undergo mechanical loading that induces stress and potential damage. The microstructural characteristics of CFRC significantly influence their mechanical behavior and durability under such conditions. Therefore, understanding the interplay between microstructural features and deformation mechanisms is critical for accurately assessing their performance and enhancing reliability in demanding environments.

Finite Element Methods (FEM)~\cite{FEM} revolutionized structural analysis by providing approximate solutions to complex differential equations governing physical phenomena. FEM is widely used in analyzing stresses, strains, and deformations in structures, optimizing designs, and predicting performance with high accuracy. Within the field of CFRC, FEM simulations have long been a primary tool for investigating deformation and damage. However, their application to laminate composites is often hindered by computational scalability challenges. These composites’ intricate geometries and material compositions necessitate the use of fine meshes, which significantly increases computational time. Although advancements such as the interface-enriched generalized FEM (IGFEM), developed by Soghrati~\cite{IGFEM}, have improved efficiency by eliminating the need for conforming meshes, FEM simulations remain computationally intensive compared to surrogate data-driven models~\cite{MGN}.

Deep learning (DL) has emerged as a powerful alternative to traditional numerical methods, demonstrating success in fields such as computational fluid dynamics~\cite{FAN, Liu, wang2024, du2024conditional}, heat transfer~\cite{Li, akharediffhybrid, kim2024}, and computational mechanics~\cite{CROOM2022104191}. Specifically, in computational mechanics, DL has been widely adopted to aid manufacturing~\cite{CVI, PINDIFF,Luo_additive, Michopoulos}, predict composite microstructural response ~\cite{Maurizi, yacouti_integrated_2025, khorrami, Yacouti}, and identify damage mechanisms~\cite{SEPASDAR_deep_learning, YAN2024110278, xu}. U-Net, a type of Convolutional Neural Networks (CNNs), has emerged as a leading architecture for precise localization and segmentation tasks, originally excelling in medical imaging~\cite{UNET}. Its symmetric encoder-decoder structure, enhanced with skip connections, captures contextual information and reconstructs high-resolution spatial features, making it particularly effective in computational mechanics and damage modeling.

Building on these advancements, neural network models have been developed to address challenges in predicting the mechanical behavior of CFRC. Sepasdar et al. employed a U-Net-based model to predict stress fields during early damage initiation and the final damage state, leveraging the architecture’s spatial feature-capturing capabilities~\cite{SEPASDAR_deep_learning}. However, this model lacked the ability to simulate the deformation process, thus failing to provide insights into the progression of material response under applied load. Xu et al. extended this approach by predicting both damage fields and stress-strain behavior throughout deformation, providing a dual perspective~\cite{xu}. Nonetheless, their model fell short in resolving micro-level stress fields, which are critical for identifying high-stress regions prone to failure. 

Despite these contributions, no existing model simultaneously predicts both stress and damage fields throughout the entire deformation process that integrates macro- and micro-level phenomena. To bridge these gaps, we propose Composite-Net, a DL approach utilizing a U-Net architecture to concurrently predict stress and damage fields across the deformation process. Composite-Net leverages U-Net’s strengths in spatial feature extraction and integrates macro-scale mechanical responses with microstructural insights. Our results demonstrate that Composite-Net achieves a root mean squared error (RMSE) of less than 15~MPa in stress distribution and an RMSE of less than 40\% in damage in over 90\% of the cases. Additionally, the model significantly outperforms IGFEM in computational efficiency, reducing per-case runtime from 500 seconds to 8 seconds, achieving a speed-up of over 60 times. By providing a unified framework for predicting CFRC behavior, the proposed deep-learning-based model in this work aims to advance the understanding of deformation and failure mechanisms, offering a powerful tool for optimizing material performance and reliability in advanced engineering applications.


\section{Methodology}
\subsection{Overview of FEM Simulation Setup}
\begin{figure}[h]
\includegraphics[width=\linewidth]{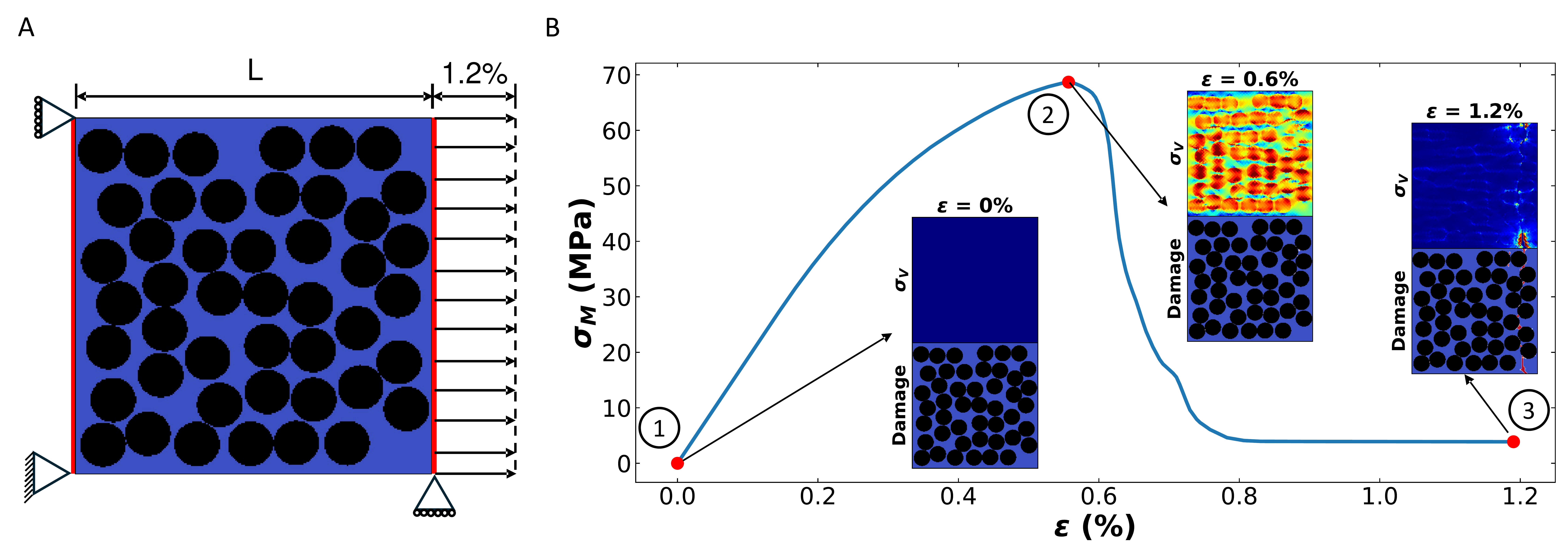}
\centering
\caption{(A) The schematic of a microstructural representation of the composite under uniaxial applied displacement with the applied boundary conditions. (B) The obtained stress-strain response: The initial state (1), the UTS (2) indicating peak material strength, and the final stage representing material failure (3). The contours present the von Mises stress and damage distribution within the microstructural representation of the composite at different strain levels. \cite{SEPASDAR_FEM}}
\label{fig:microstructure_stress_strain}
\end{figure}

The composite structure under investigation, shown in Figure~\ref{fig:microstructure_stress_strain}-A, is a two-dimensional (2D) unidirectional CFRC comprising carbon fibers oriented out of the plane (90$^\circ$ ply). The CFRC is subjected to a displacement-controlled uniaxial tensile test, where strain is incrementally applied up to 1.2\% to examine its mechanical response under progressive deformation, including failure. The vertical displacement is fixed at the bottom edge of the microstructure, while the left edge is constrained to have no horizontal displacement. On the right edge, a Dirichlet boundary condition is imposed, where displacement is incrementally applied using a pseudo-time-stepping methodology to ensure numerical stability and accurately track damage progression. Fibers are embedded within an epoxy matrix, and the interactions between the two phases are explicitly modeled through a cohesive zone model. The carbon fibers have a transversely isotropic elastic response, while the epoxy matrix is modeled using an elasto-plastic damage model to capture the heterogeneous stress distribution and damage evolution.

The mechanical behavior of the CFRC under the applied displacement is summarized by the stress-strain curve shown in Figure~\ref{fig:microstructure_stress_strain}-B. Key points on the curve include the initial state (1), the ultimate tensile stress (UTS) corresponding to peak material strength (2), and the final region representing material failure (3), highlighting the stress and damage evolution under the specified applied displacement. Additional information on the FEM simulations, including a detailed description of the constitutive equations and the material properties, is also included in~\ref{appendix:FEM_details} to provide further clarity. FEM simulations serve as the ground truth for training and validating the surrogate model developed in this study.

\subsection[h]{Data Generation and Preprocessing}
The random fiber generation algorithm used in this study was adapted from the method developed in \cite{SEPASDAR_deep_learning}. This algorithm ensures that the generated fiber locations maintain a specified fiber volume fraction while adhering to a defined nearest-neighbor distance (NND) distribution. The ability of this approach to capture the variability in microstructures has been previously validated in \cite{SEPASDAR_deep_learning}. For the current analysis, microstructures were generated with square dimensions of \( 54 \, \mu\text{m} \times 54 \, \mu\text{m} \), each containing circular fibers with a diameter of \( 7 \, \mu\text{m} \). Using the algorithm, a total of 2000 distinct microstructures were generated and analyzed using the IGFEM framework to produce high-fidelity training and testing data, with data augmentation applied by mirroring the microstructures along the y-axis, thereby doubling the dataset size. It is important to emphasize that the parameters for the constitutive equations remained constant across all simulations, with the only variable being the spatial arrangement of the fibers.

The microstructure, von Mises stress, component stress, and damage generated through IGFEM are in the form of an unstructured mesh in Visualization Toolkit (VTK) format. A data processing pipeline utilizing PyVista ~\cite{pyvista} is employed to convert an unstructured finite element mesh into a structured 256~$\times$~256 grid suitable for Composite-Net. The unstructured mesh, which consists of irregularly spaced nodes and elements, is first loaded into PyVista as an unstructuredGrid object. An uniform Cartesian grid of size 256~$\times$~256 is then generated to define a structured domain. Microstructure, von Mises stress, component stress, and damage are interpolated from the unstructured mesh to the structured grid using PyVista's sample method, which employs point-wise interpolation based on inverse distance weighting or finite element shape functions. To mitigate interpolation artifacts and ensure numerical accuracy, integration-based averaging is applied when mapping element-based quantities. This transformation facilitates the use of CNN by providing a consistent grid-based input while preserving the fidelity of the original FEM solution.

In order to ensure that the input data is appropriately scaled, a normalization procedure is applied. Specifically, each feature is normalized by subtracting its mean and dividing by its standard deviation, as shown in the following equation:

\begin{equation}
\mathbf{x}_{\text{norm}} = \frac{\mathbf{x} - \mu}{std}
\label{eq:normalization}
\end{equation}

where \(\mathbf{x}\) represents the raw data, \(\mu\) is the mean, and \(std\) is the standard deviation of the data. This standardization ensures that all input features have zero mean and unit variance, allowing the model to better converge during training. The normalization procedure is applied to each of the features: microstructure, von Mises stress, stress components (i.e., $\sigma_{11}$, $\sigma_{22}$, and $\sigma_{12}$), and damage fields, etc. By normalizing the data, we reduce the impact of scale differences among the various features and enhance the model's ability to learn the underlying patterns in the data.


\subsection[h]{Deep Learning Framework}
\begin{figure}[h]
\includegraphics[width=\linewidth]{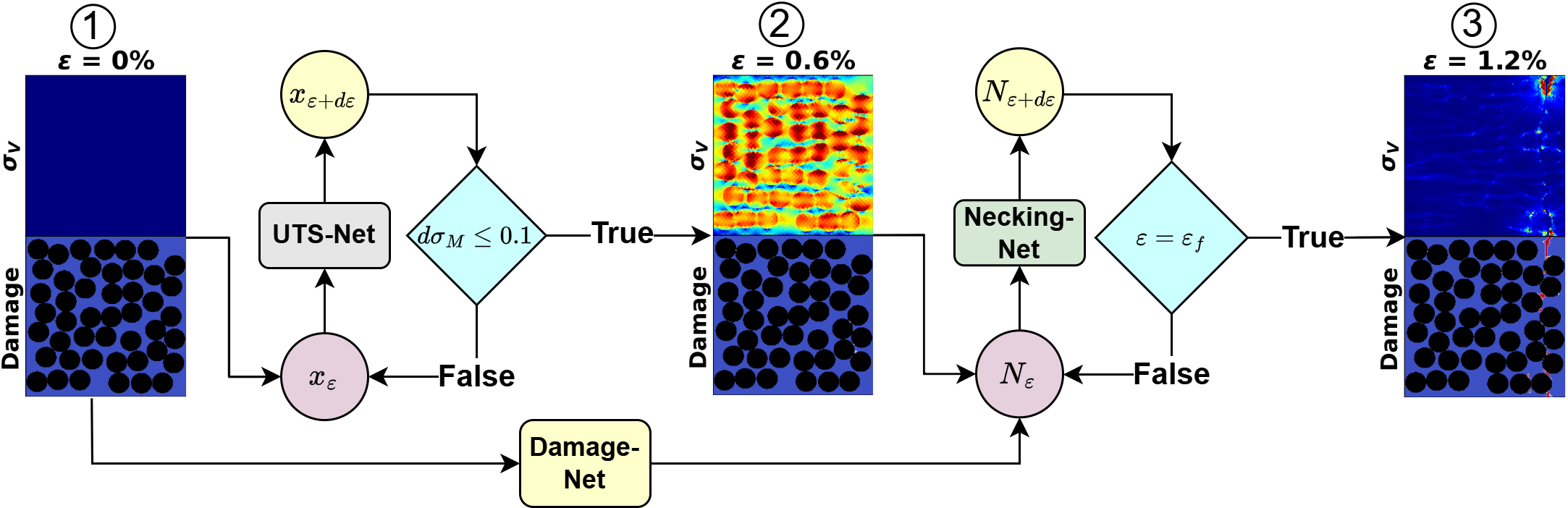}
\centering
\caption{Schematic representation of Composite-Net, consisting of three interconnected neural networks: Damage-Net (yellow) for mapping microstructure to the final damage pattern, UTS-Net (gray) for predicting the response from initial condition to UTS, and Necking-Net (green) for modeling the transition from UTS to $\epsilon_f$. $x_{\epsilon}$ and $N_{\epsilon}$ are the input features of the UTS-Net and Necking-Net, respectively; $x_{\epsilon+d\epsilon}$ and $N_{\epsilon+d\epsilon}$ are the output features of the UTS-Net and Necking-Net, respectively;}
\label{fig:model_structure}
\end{figure}

The DL framework proposed in this work, Composite-Net, is inspired by ensemble learning~\cite{Ensemble}, which integrates multiple neural networks as shown in Figure~\ref{fig:model_structure} to model the FEM simulation process, addressing the limitations of using a single DL model for the entire prediction. A single DL model struggles to capture the full material behavior due to evolving composite properties, such as the stiffening or softening of the matrix, which are not explicitly provided in the DL model. To overcome such limitations and predict the composite's microscopic (i.e., stress and damage distribution within the microstructural representation of the CFRC) responses throughout the loading history, Composite-Net is divided into three specialized components: 
\begin{itemize}
    \item \textbf{Damage-Net}, which maps the microstructure to the final damage pattern.
    \item \textbf{UTS-Net}, which predicts spatiotemporal damage and stress from the initial condition to UTS.
    \item \textbf{Necking-Net}, which models the spatiotemporal damage and stress from UTS to the final strain, $\epsilon_f$ (i.e., $\epsilon = 1.2\%$ in this work because that is when the FEM simulation ends), focusing on the necking behavior that leads to failure.
\end{itemize}
Composite-Net is designed to switch from UTS-Net to Necking-Net by checking if the increase of macro stress, $d\sigma_M$, is less than 0.1~MPa, because an increase of less than 0.1~MPa is considered negligible when the macro stress is more than 60~MPa. Necking-Net will stop when the applied strain, $\epsilon$, reaches 1.2\%. This composite neural network design enables a more accurate and systematic prediction of both microstructural and macroscopic responses of CFRC across different stages of the deformation process. The details of each neural network will be further described in the following sections.

The dataset used to train Damage-Net consists of 2000 unique microstructures and their corresponding final damage patterns. In contrast, the dataset used to train UTS-Net and Necking-Net contains only 200 unique cases. By incorporating sequential time steps capturing the full deformation process, rather than just initial and final states, the models effectively learn from intermediate stages, reducing the number of unique cases required for training.

\subsubsection{U-Net Architecture}
\begin{figure}[h]
\includegraphics[width=\linewidth]{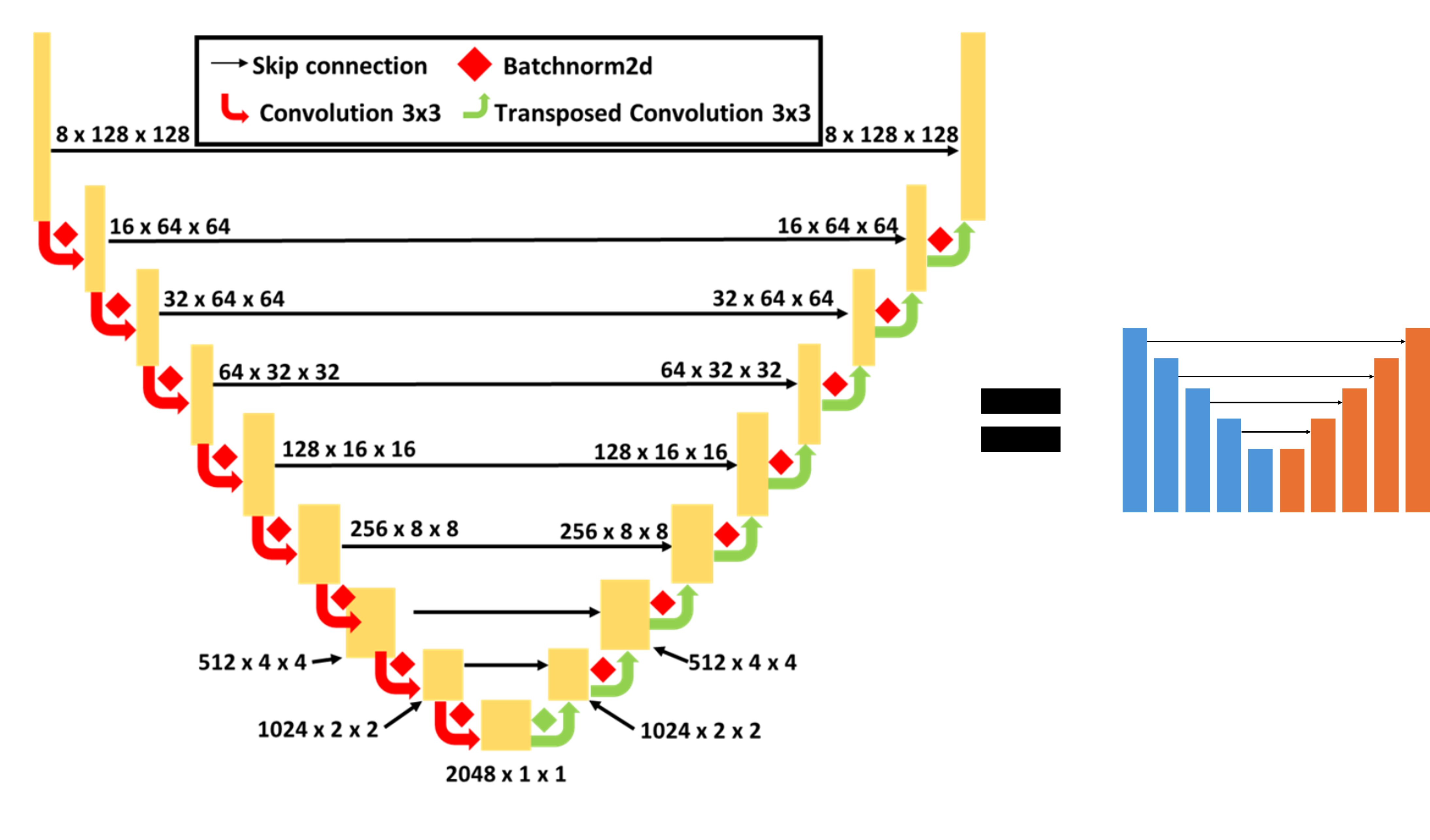}
\centering
\caption{Architecture of U-Net used in Damage-Net, UTS-Net and Necking-Net (Left). The simplified schematic of the U-Net architecture (Right).}
\label{fig:U-Net_architecture}
\end{figure}

The U-Net architecture and the simplified schematic used in Damage-Net, UTS-Net, and Necking-Net are illustrated in Figure~\ref{fig:U-Net_architecture}. Damage-Net, UTS-Net, and Necking-Net use the same U-Net architecture. The network contains two types of convolutional layers: Convolution and Transposed Convolution, which encode and decode the input features, respectively. The Convolution layer transforms the input of dimension c~$\times$~256~$\times$~256, where c is the number of features in the input, to 8~$\times$~256~$\times$~256, and then into a lower-dimensional 2048~$\times$~1~$\times$~1 vector of features that describes the input image. Next, the Transposed Convolution translates the encrypted information in the 2048~ $\times$~1~$\times$~1 vector and transforms the vector into outputs with dimension c~$\times$~256~$\times$~256. 

At each block of both the encoder and decoder, batch normalization is applied to stabilize training, improve convergence speed, and mitigate internal covariate shift. Additionally, skip connections are incorporated between all corresponding blocks of the encoder and decoder to facilitate learning and generation by enhancing information flow and preserving spatial features.

\subsubsection{Damage-Net}
\begin{figure}[h]
\includegraphics[width=\linewidth]{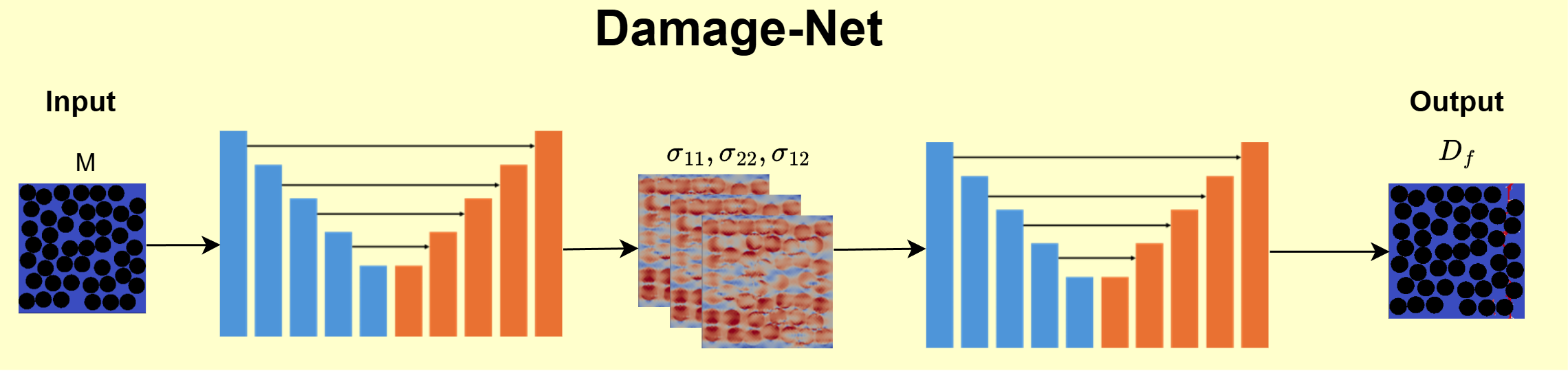}
\centering
\caption{Schematic representation of the Damage-Net for mapping microstructure to the final damage pattern.}
\label{fig:dmg-net}
\end{figure}

Damage-Net is a dual U-Net-based deep learning framework inspired by Sepasdar~\cite{SEPASDAR_deep_learning}, designed to predict the final damage patterns within composite microstructures, as shown in Figure~\ref{fig:dmg-net}. The architecture of Damage-Net consists of two sequentially connected U-Net models. The first U-Net model predicts the contours of the stress components: stress in x-direction, $\sigma_{11}$, stress in y-direction, $\sigma_{22}$, and shear stress, $\sigma_{12}$ at UTS, identifying critical regions within the microstructure where damage is likely to initiate under applied loading conditions. The second U-Net model utilizes the predicted $\sigma_{11}$, $\sigma_{22}$, and $\sigma_{12}$ contours as an intermediate state to map the microstructure to the final damage pattern.

The first U-Net is trained using a hybrid loss function that combines a physics-informed loss and the Mean Squared Error (MSE) loss. This ensures that the model captures both data and physical consistency, resulting in more accurate and realistic predictions of $\sigma_{11}$, $\sigma_{22}$, and $\sigma_{12}$. The MSE loss is defined as
\begin{equation}
\label{eq:mse_dmg}
\mathcal{L}_{\text{MSE}} = \frac{1}{N} \sum_{n=1}^{N} \frac{1}{P_n} \sum_{p=1}^{P_n} \frac{1}{3}\sum_{c \in \{\sigma_{11}, \sigma_{22}, \sigma_{12}\}} \left( c^{(n, p)} - \hat{c}^{(n, p)} \right)^2
\end{equation}
where \( N \) is the total number of training cases, \( P_n \) is the number of pixels in the contour for the \( n \)-th case, and \( c \) represents the stress components \( \sigma_{11}, \sigma_{22}, \) and \( \sigma_{12} \). The terms \( c^{(n, p)} \) and \( \hat{c}^{(n, p)} \) correspond to the ground truth and predicted features, respectively. The loss is averaged over all pixels within the contour and across all training cases to ensure robust learning of stress field predictions.

In addition to the MSE loss, a physics-based equilibrium residual loss enforces that the predicted $\sigma_{11}$, $\sigma_{22}$, and $\sigma_{12}$ satisfy the stress equilibrium equation
\begin{equation}
\nabla \cdot \boldsymbol{\sigma} = 0,
\end{equation}
where $\boldsymbol{\sigma}$ is the stress tensor. This residual loss is expressed as
\begin{equation}
\mathcal{L}_\text{physics} = \frac{1}{N} \sum_{n=1}^{N} \frac{1}{P_n} \sum_{p=1}^{P_n} \| \nabla \cdot \boldsymbol{\hat{\sigma}}^{(n, p)} - \nabla \cdot \boldsymbol{\hat{\sigma}}^{(n, p)} \|^2.
\end{equation}

The total loss function for the first U-Net is the average of the two components
\begin{equation}
\mathcal{L}_\text{total} = 0.5 \mathcal{L}_\text{MSE} + 0.5 \mathcal{L}_\text{physics},
\end{equation}
This approach ensures that the $\sigma_{11}$, $\sigma_{22}$, and $\sigma_{12}$ predictions are both accurate and aligned with the fundamental principles of mechanics.

The second U-Net, tasked with predicting the final damage pattern, is trained using the Binary Cross-Entropy (BCE) loss, which is suitable for damage recognition given its binary state (i.e., either damaged or undamaged). The BCE loss is given by Eq.~\eqref{eq:bce}. \(y_{n,p}\) represents the ground truth damage at pixel \( p \) in the \( n \)-th training case, indicating the material is damaged (\(y_i = 1\)) or not damage (\(y_i = 0\)). \(\hat{y}_{n,p}\) is the predicted damage parameter at pixel \( p \) in the \( n \)-th training case. 

\begin{equation}
\label{eq:bce}
\mathcal{L}_\text{BCE} = \frac{1}{N} \sum_{n=1}^{N} \left(-\frac{1}{P_n}\sum_{i=1}^{P_n} 
\left[ y_{n,p}\cdot \log(\hat{y}_{n,p}) + (1 - y_{n,p})\cdot \log(1 - \hat{y}_{n,p}) \right]\right).
\end{equation}

The BCE loss penalizes incorrect predictions based on their confidence, encouraging the model to accurately distinguish between damaged and undamaged regions. This probabilistic approach enables the second U-Net to model the complex relationship between the $\sigma_{11}$, $\sigma_{22}$, and $\sigma_{12}$ fields at UTS and the damage distribution, providing a robust foundation for predicting damage initiation and propagation in CFRC. $\sigma_{11}$, $\sigma_{22}$, and $\sigma_{12}$ are used in Damage-Net instead of simply using von Mises stress at UTS because they are needed to calculate the physics-based stress-equilibrium residual loss.

\subsubsection{UTS-Net}
\begin{figure}[h]
\includegraphics[width=\linewidth]{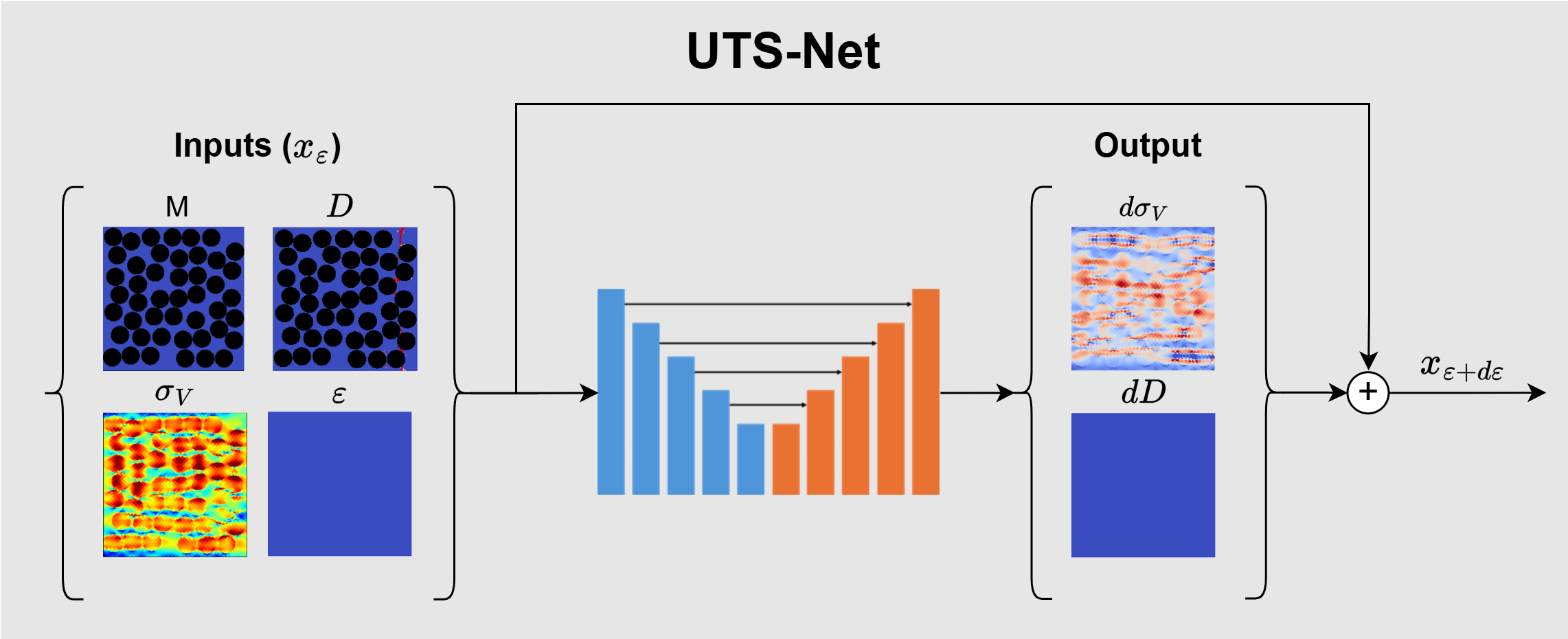}
\caption{Schematic representation of UTS-Net for predicting the von Mises stress and damage response from initial condition to UTS.}
\label{fig:UTS-net}
\end{figure}

UTS-Net predicts the spatiotemporal von Mises stress and damage from the initial state up to UTS. The network takes as input the microstructure (M), applied strain ($\epsilon$), von Mises stress ($\sigma_v$), and damage state ($D$) at a given strain, as shown in Figure~\ref{fig:UTS-net}. The $\epsilon$ contour is the $\epsilon$ value replicated across the contour, matching the dimensions of the other inputs, to integrate it as part of the input for UTS-Net. Using these inputs, UTS-Net predicts d$\sigma_V$ and d$D$, the differences in $\sigma_V$ and $D$ relative to the input state, respectively. These predicted differences are then added to the input fields to calculate $\sigma_V$ and $D$ at the next applied strain, $\epsilon + d\epsilon$. The auto-regressive process is iterated until $d\sigma_M$, the pixel average of $d\sigma_V$, is less than 0.1~MPa. By leveraging microstructural information and focusing on incremental updates, UTS-Net effectively captures the evolution of stress and damage during the pre-UTS stage, providing a foundation for modeling subsequent phases of deformation.

UTS-Net is trained using MSE loss in Eq.~\eqref{eq:mse_UTS}. The terms \( X^{(n, p, t)} \) and \( \hat{X}^{(n, p, t)} \) correspond to the ground truth and predicted values, respectively, for feature \( X \) at pixel \( p \) at applied strain \( t \) in the \( n \)-th training case. The model is trained using ground truth d$\sigma_V$ and d$D$, because the neural network has better stability when predicting the change at each $\epsilon$ than predicting $\sigma_V$ and $D$ at each $\epsilon$. UTS-Net does not use BCE loss for training, because the damage field represents a continuously evolving scalar quantity rather than a binary classification. The model predicts incremental changes in stress and damage at each applied strain, which are added to the input fields to compute updates for the next step. This continuous evolution aligns better with the MSE shown in Eq.~\eqref{eq:mse_UTS}, which emphasizes accurate scalar field predictions, rather than BCE, which is suited for binary or localized tasks. Therefore, UTS-Net is trained using MSE rather than BCE.

\begin{equation}
\label{eq:mse_UTS}
\mathcal{L}_{\text{MSE}} = \frac{1}{N} \sum_{n=1}^{N} \frac{1}{T} \sum_{t=1}^{T} \frac{1}{P_n} \sum_{p=1}^{P_n} \frac{1}{2}\sum_{X \in \{d\sigma_V, dD\}} \left(X^{(n, p, t)} - \hat{X}^{(n, p, t)} \right)^2
\end{equation}

\subsubsection{Necking-Net}
\begin{figure}[h]
\includegraphics[width=\linewidth]{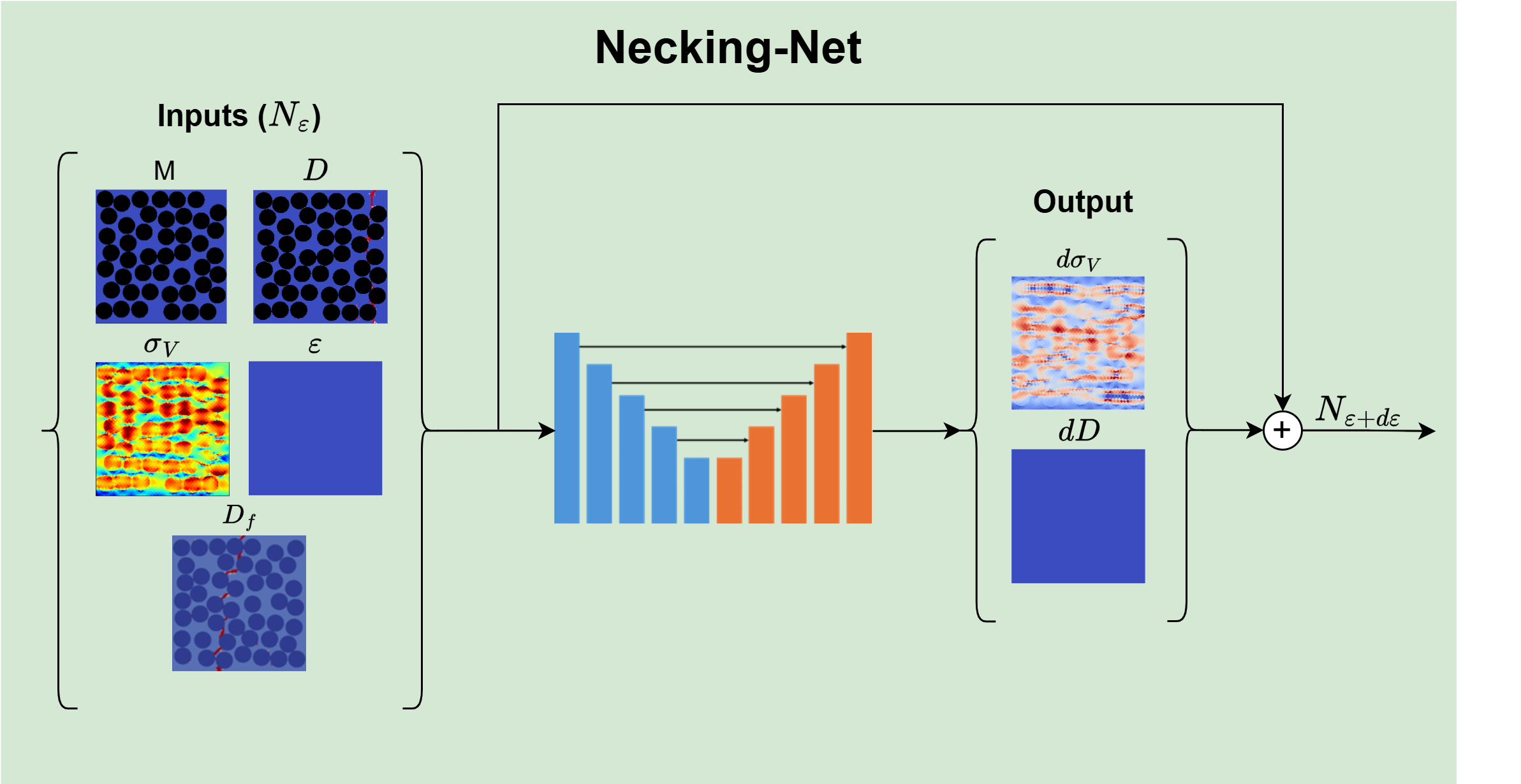}
\centering
\caption{Schematic representation of Necking-Net for predicting the von Mises stress and damage response from UTS to final strain.}
\label{fig:Necking-net}
\end{figure}

Necking-Net predicts the evolution of stress and damage fields during the critical transition from UTS to the failure strain. The network takes as input the microstructure (M), applied strain ($\epsilon$), von Mises stress ($\sigma_V$), and damage state ($D$), and the final damage state ($D_{f}$), as shown in Figure~\ref{fig:Necking-net}. The $\epsilon$ contour is the $\epsilon$ value replicated across the contour, matching the dimensions of the other inputs, to integrate it as part of the input for UTS-Net. Using these inputs, Necking-Net predicts d$\sigma_V$ and d$D$, the differences in $\sigma_V$ and $D$ relative to the input state, respectively. These predicted differences are added to the input fields to compute $\sigma_V$ and $D$ at $\epsilon + d\epsilon$. This iterative process is rolled out until the condition $\epsilon = \epsilon_{\text{f}}$ is met. By leveraging detailed microstructural information and capturing the complex redistribution of stresses and damage evolution during necking, Necking-Net provides accurate predictions of material behavior until final strain. 

Necking-Net employs MSE loss shown in Eq.~\eqref{eq:mse_UTS} because the model also predicts continuous changes in the necking profile, which evolves gradually rather than as discrete or binary events. 


\section{Results}
The results section is organized into four key parts to systematically evaluate the proposed DL models. The first three subsections focus on Damage-Net, UTS-Net, and Necking-Net, respectively, discussing their individual contributions to predicting final damage, evolution of the von Mises stress and damage fields up to UTS, and post-UTS behavior. The fourth section integrates these models into Composite-Net, a composite framework, demonstrating their combined effectiveness in comprehensively capturing the deformation and failure mechanisms in CFRC. All four models are evaluated using the same testing dataset, which comprises 100 unseen cases. RMSE is employed as the evaluation metric. RMSE measures the average magnitude of errors between the predictions and ground truth, offering a comprehensive assessment of prediction precision.

\begin{figure}[h]
\includegraphics[width=\textwidth]{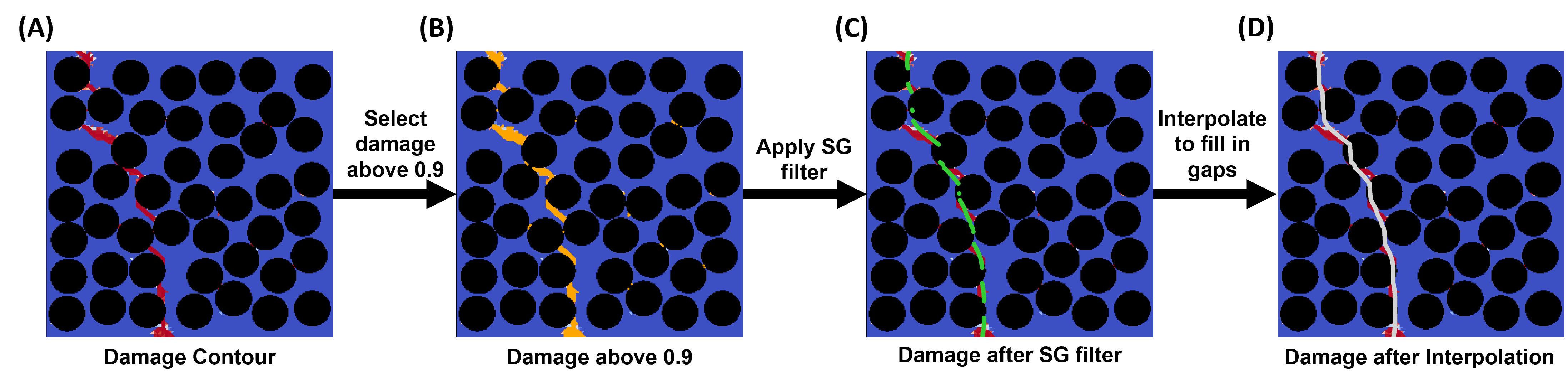}
\caption{Post-processing pipeline for extracting the main crack path from predicted damage fields. (A) Ground truth damage field; (B) Binary mask of damage pixels exceeding a threshold of 0.9; (C) Smoothed damage field after applying a Savitzky-Golay (SG) filter, with the extracted crack path highlighted in green; (D) Interpolated crack trajectory from the smoothed field to obtain a continuous and coordinate-traceable main crack path. This post-processing enables spatial tracking and quantitative comparison of predicted and reference failure trajectories.}
\label{fig:post_process}
\end{figure}

The output of Damage-Net is a 256~$\times$~256 spatial field representing damage values ranging from 0 to 1. Traditional pixel-wise metrics such as RMSE and intersection over union (IoU) are commonly used to evaluate the accuracy of damage predictions in image-based learning tasks. However, these metrics are insufficiently sensitive in the context of crack path prediction for CFRC composites. Even small spatial deviations—such as a few pixels of offset in the predicted crack trajectory—can lead to disproportionately large penalties in RMSE and IoU, making them unable to distinguish between minor localization errors and completely erroneous predictions. This lack of spatial tolerance undermines the interpretability of prediction quality in failure analysis. To address this limitation, we propose a post-processing method shown in Figure~\ref{fig:post_process} that extracts the coordinates of the main crack path and quantifies error based on deviations in the x-coordinate along the vertical axis. This coordinate-based evaluation provides a more physically meaningful measure of crack path accuracy, enabling robust comparison between predicted and ground-truth failure trajectories.

The method is applied to both the predicted and ground truth damage fields. First, the raw damage field (Figure~\ref{fig:post_process}A) is filtered to produce a binary mask, where pixels with damage values above 0.9 are retained (Figure~\ref{fig:post_process}B). This step isolates the significantly damaged regions. Next, the horizontal (i.e., column-wise) damage indices are extracted for each vertical index (i.e., row) where damage occurs, converting the 2D damage map into a 1D representation of the main crack trajectory. A median filter is applied to the damage indices to reduce noise and remove outliers, followed by a Savitzky-Golay (SG) filter to smooth the path (Figure~\ref{fig:post_process}C). The smoothed damage path is then interpolated across all vertical indices to generate a continuous line representation (Figure~\ref{fig:post_process}D).

To quantify prediction precision, \% RMSE is calculated between the horizontal indices of the predicted crack paths and label crack paths at each corresponding vertical index using Eq.~\eqref{eq:mse_D}. This index-based error metric captures the deviation in crack location across the field, providing a more robust and interpretable assessment of prediction quality than conventional pixel-wise RMSE or IoU, which fail to distinguish between small spatial shifts and entirely incorrect predictions.

\begin{equation}
\label{eq:mse_D}
\text{\% RMSE} = \sqrt{\frac{1}{L} \sum_{i=1}^{L}  \left(\frac{X^i - \hat{X^i}}{256} \right)^2}\times 100\%
\end{equation}
\( L \) is the total number of pixels in the y-direction. $X^i$ and $\hat{X}^i$ represent the ground truth and predicted x-index of the main crack at the y-index \(i\), respectively.

To quantify the accuracy of the von Mises stress predictions, the RMSE is computed as follows
\begin{equation}
\text{RMSE}_{\sigma_V} = \sqrt{\frac{1}{T} \sum_{t=1}^T\frac{1}{P} \sum_{p=1}^P (\sigma_V^{t,p} - \hat{\sigma}_V^{t,p})^2},
\end{equation}
where \(\sigma_V^{t,p} \) and \(\sigma_V^{t,p} \) represents the von Mises stress at applied strain \(t\) and pixel \(p\). Lower RMSE values indicate closer agreement between the predicted and true stress fields, while higher values highlight discrepancies. In the following subsections, examples of the predictions of the best, average, and worst cases of each model will be presented. The average von Mises stress in the average von Mises stress vs. strain curve is computed using Eq.~\eqref{eq:stress_M}. 

\begin{equation}
\label{eq:stress_M}
\sigma_V = \frac{1}{P} \sum_{p=1}^P \sigma_V^{p},
\end{equation}

\subsection{Damage-Net}
\begin{figure}[h]
\includegraphics[width=\textwidth]{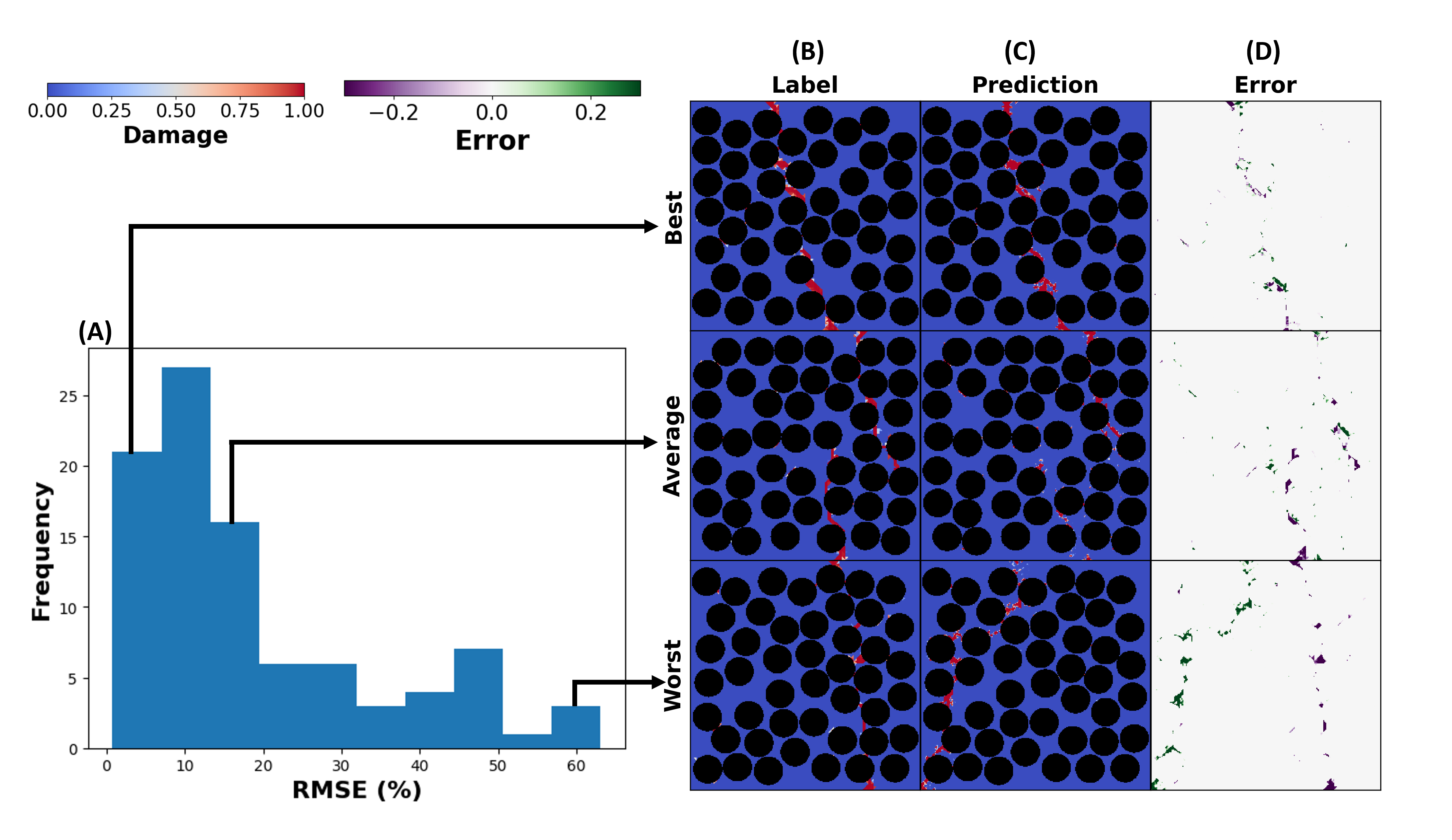}
\caption{Histogram of RMSE values for the testing dataset using Damage-Net (A). Column B shows the ground truth damage patterns. Column C presents the Damage-Net predictions. Column D illustrates the error distributions.}
\label{fig:damage_net_results}
\end{figure}

Figure~\ref{fig:damage_net_results} presents the results of the Damage-Net predictions on the testing dataset, highlighting the best, average, and worst cases based on RMSE. The figure consists of three columns: the Column B shows the ground truth final damage patterns, Column C displays the corresponding model predictions, and Column D visualizes the spatial error.

For the best and average cases, the Damage-Net predictions closely replicate the ground truth damage patterns, with an RMSE ranging from 0 to 20\%. This range reflects the model's ability to accurately capture the final damage pattern. The predictions of stress components from the first U-Net provide a reliable foundation for the second U-Net, enabling accurate prediction of the final damage pattern. Specifically, in the testing dataset, over 50\% of the cases have an RMSE of less than 20\%, demonstrating the model's capacity to predict the final damage pattern with high accuracy. In the worst-case scenario, discrepancies in the final damage pattern are observed; however, these discrepancies affect only less than 5\% of the testing dataset, further illustrating the model's robustness. Overall, the results demonstrate the effectiveness of Damage-Net in predicting the final damage pattern from microstructure.

\subsection{UTS-Net}
\begin{figure}[h]
\includegraphics[width=\textwidth]{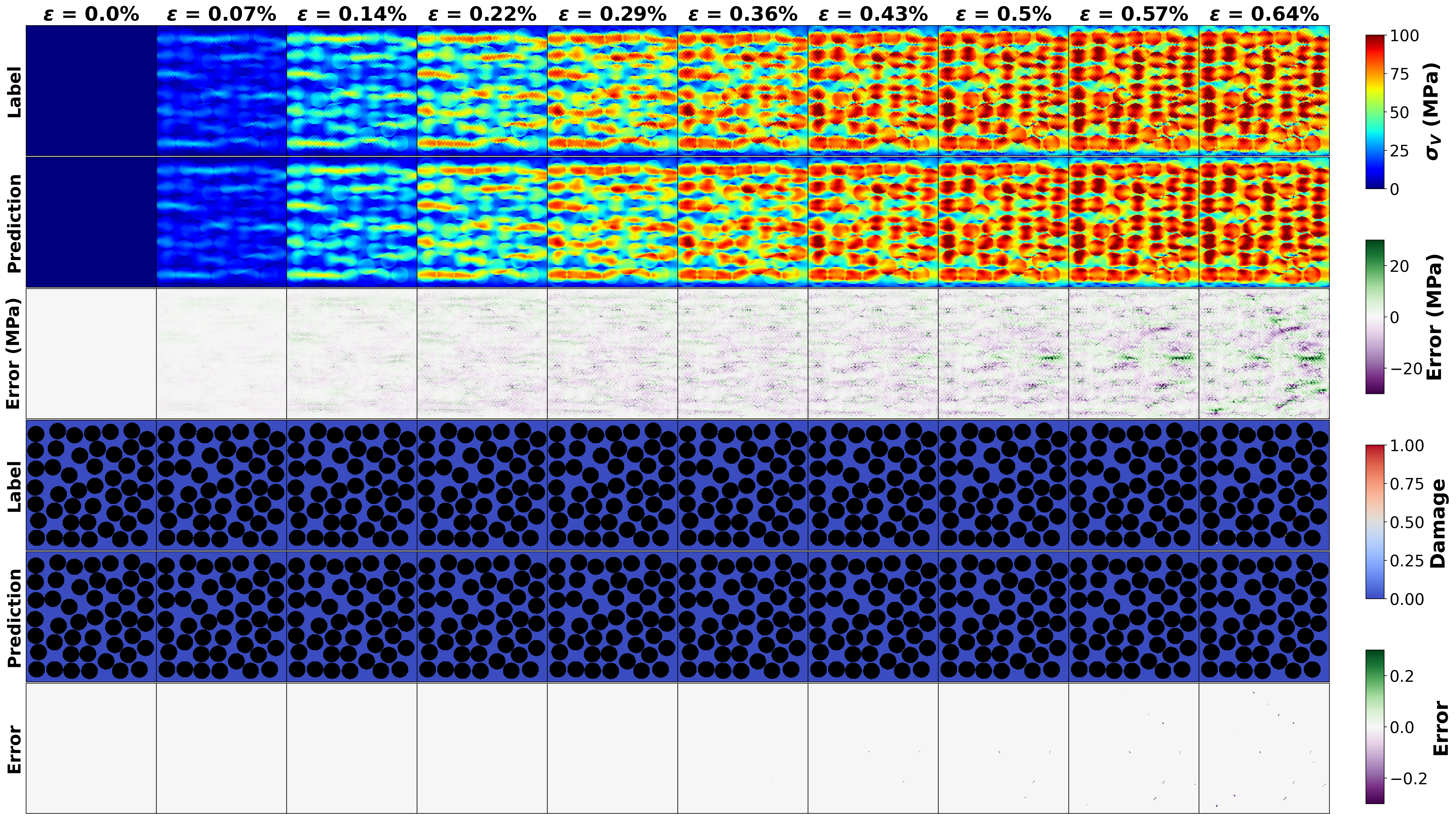}
\centering
\caption{Rollout predictions and ground truth of stress (top) and damage (bottom) fields from UTS-Net.}
\label{fig:UTS_t}
\end{figure}

Figure~\ref{fig:UTS_t} illustrates the rollout predictions for a representative test case, showing the model's ability to predict the evolving von Mises stress and damage fields over successive strain increments up to UTS. The rollout damage predictions before reaching UTS show no significant damage accumulation, consistent with the expected material behavior where damage mechanisms are minimal or inactive prior to UTS. The error plots also show minimum error accumulation in both damage and stress fields, proving UTS-Net's stability to accurately perform predictions prior to UTS. The rollout predictions demonstrate UTS-Net's capability to accurately capture the spatiotemporal stress and damage until UTS and prove its capability as the first part model of Composite-Net.

\begin{figure}[h]
\includegraphics[width=\textwidth]{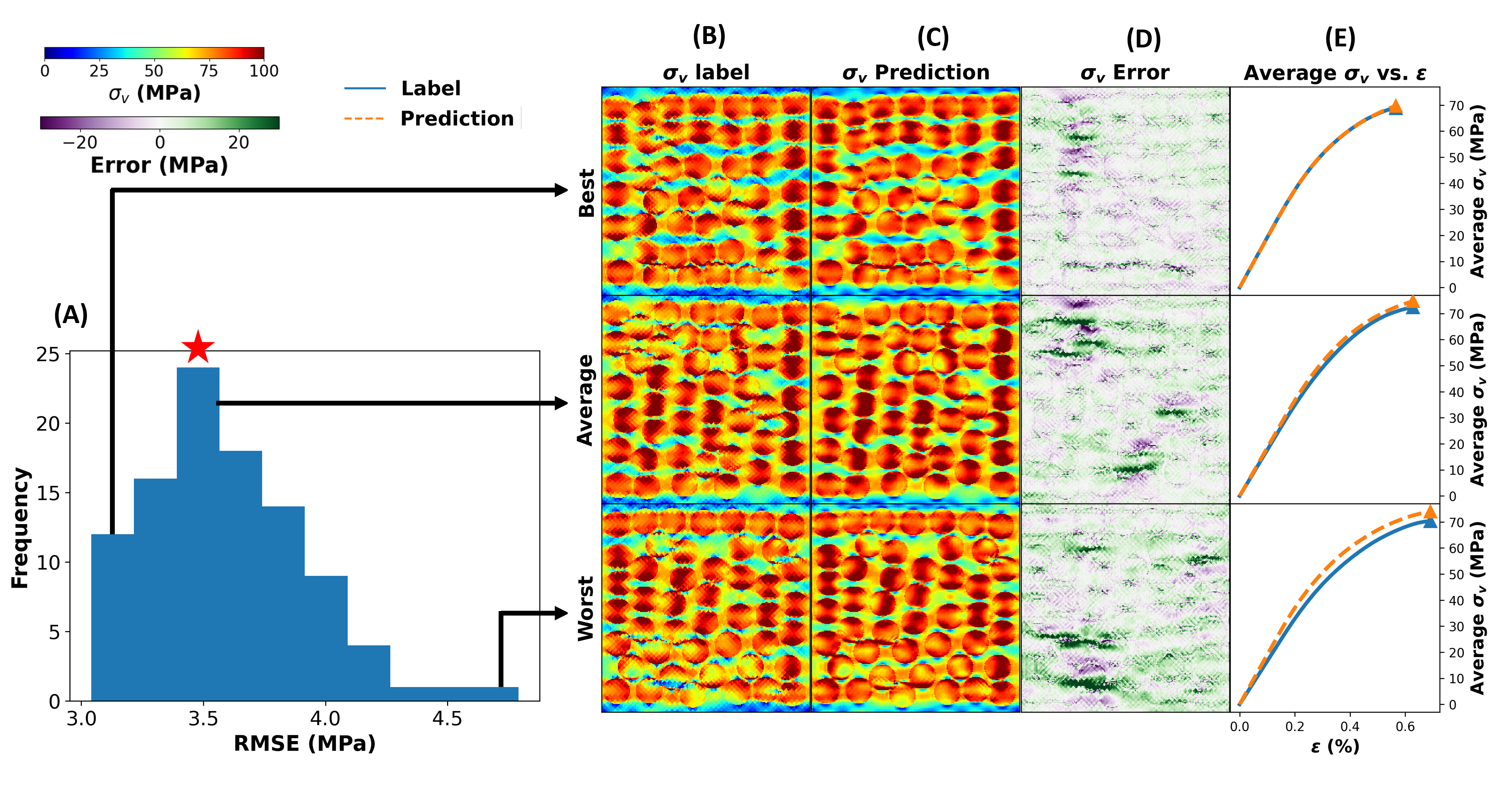}
\centering
\caption{Histogram of RMSE for the testing dataset using UTS-Net (A). The red star represents where the case in Figure~\ref{fig:UTS_t} would be in the histogram. Column B shows the stress field label data at the last time step (UTS). Column C presents the stress field predictions at UTS. Column D depicts the error distribution. Column E shows the average von Mises stress vs. strain curves. The triangles in Column E show where in the deformation process the von Mises stress contours are located. Each row in the columns corresponds to the best, average, and worst cases, determined by their respective RMSE values.}
\label{fig:UTS_result}
\end{figure}

To analyze the spatial accuracy of UTS-Net, a histogram of the RMSE of the testing dataset is plotted, along with the ground truth and predicted von Mises stress at UTS, error distributions, and average von Mises stress vs. strain curve of the best, average and worst cases, as illustrated in Figure~\ref{fig:UTS_result}. Over 50\% of the cases have RMSE values below 2.5~MPa, and all cases have RMSE values below 5~MPa, which correspond to 2.5\% and 5\% of the maximum stress magnitude, respectively. The stress field predictions agree well with the ground truth stress field. This observation is in alignment with the small errors in the error contour as well as the low RMSE values in the histogram. 
To assess the temporal prediction capabilities of UTS-Net, average von Mises stress vs. strain is plotted for the best, average, and worst cases in column E of Figure~\ref{fig:UTS_result}. The curves for the best and average cases closely follow the ground truth data, confirming the model's ability to capture the material's progressive stress response under increasing applied strain. The worst case still exhibits reasonable agreement with the ground truth. This consistency across the temporal domain demonstrates the effectiveness of the auto-regressive approach in modeling the incremental evolution of stress. 

UTS-Net is therefore well-suited for fast and accurate spatiotemporal stress and damage predictions up to the ultimate tensile stress. The results demonstrate predictions closely matching the ground truth, highlighting its robustness across varying microstructures. Overall, UTS-Net's ability to effectively predict both spatiotemporal stress and damage makes it a valuable tool for capturing the pre-UTS parts in Composite-Net.

\subsection{Necking-Net}
\begin{figure}[h]
\includegraphics[width=\textwidth]{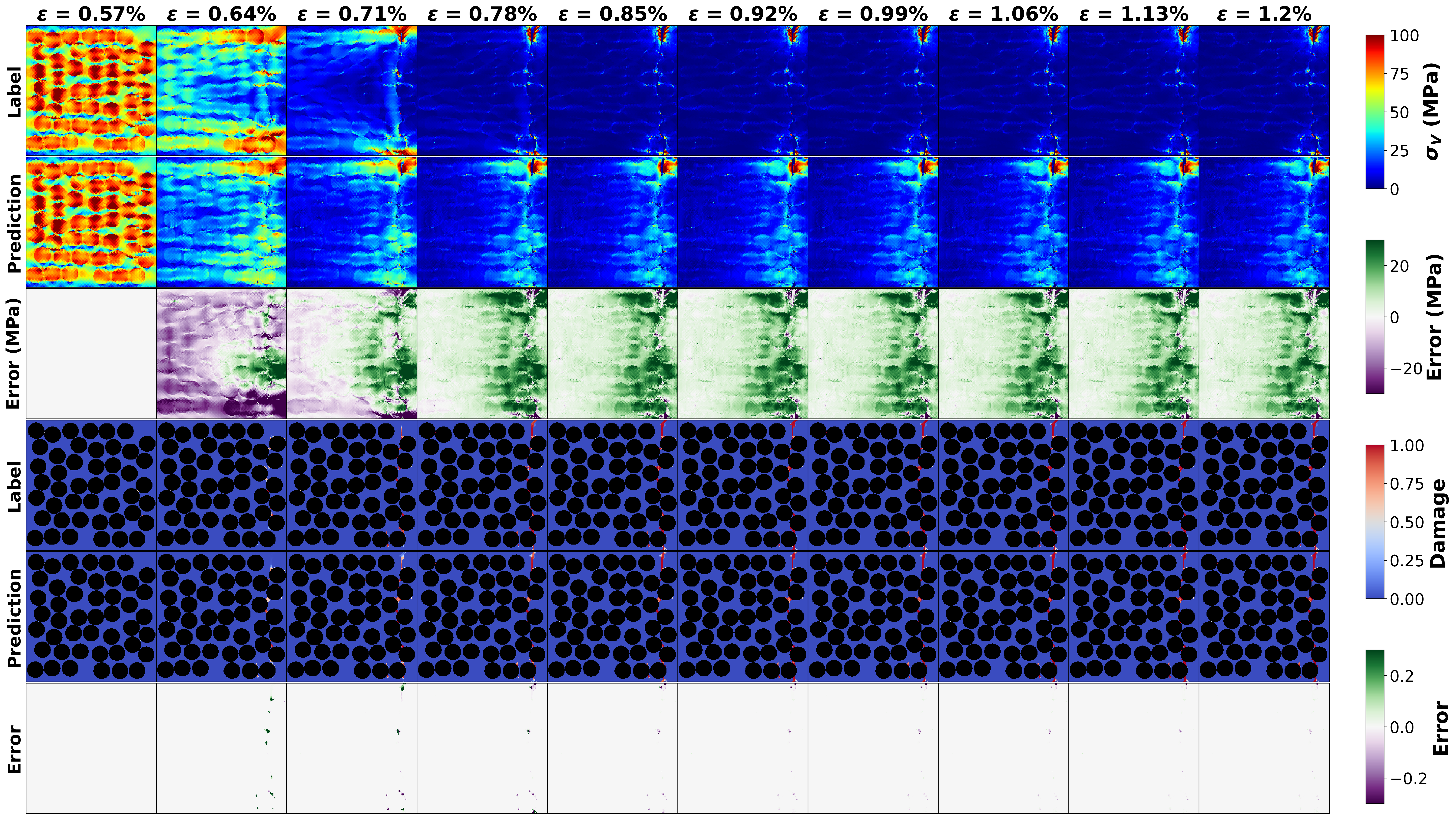}
\centering
\caption{Rollout predictions and ground truth of stress (top) and damage (bottom) fields from Necking-Net.}
\label{fig:Necking_t}
\end{figure}

Figure~\ref{fig:Necking_t} illustrates the rollout predictions for a representative test case, highlighting Necking-Net's ability to capture the progressive localization of von Mises stress and damage fields under post-UTS deformation. The rollout results demonstrate Necking-Net's effectiveness in tracking the spatial and temporal evolution of stress and damage fields post-UTS. Starting from the stress and damage fields at UTS, the model auto-regressively predicts their changes with increasing strain, capturing the concentration of stress and the increase of damage at the necking zone. The predicted fields agree well with the ground truth, replicating the spatially localized deformation patterns that characterize necking. These results show Necking-Net's spatiotemporal capability.

\begin{figure}[h]
\includegraphics[width=\textwidth]{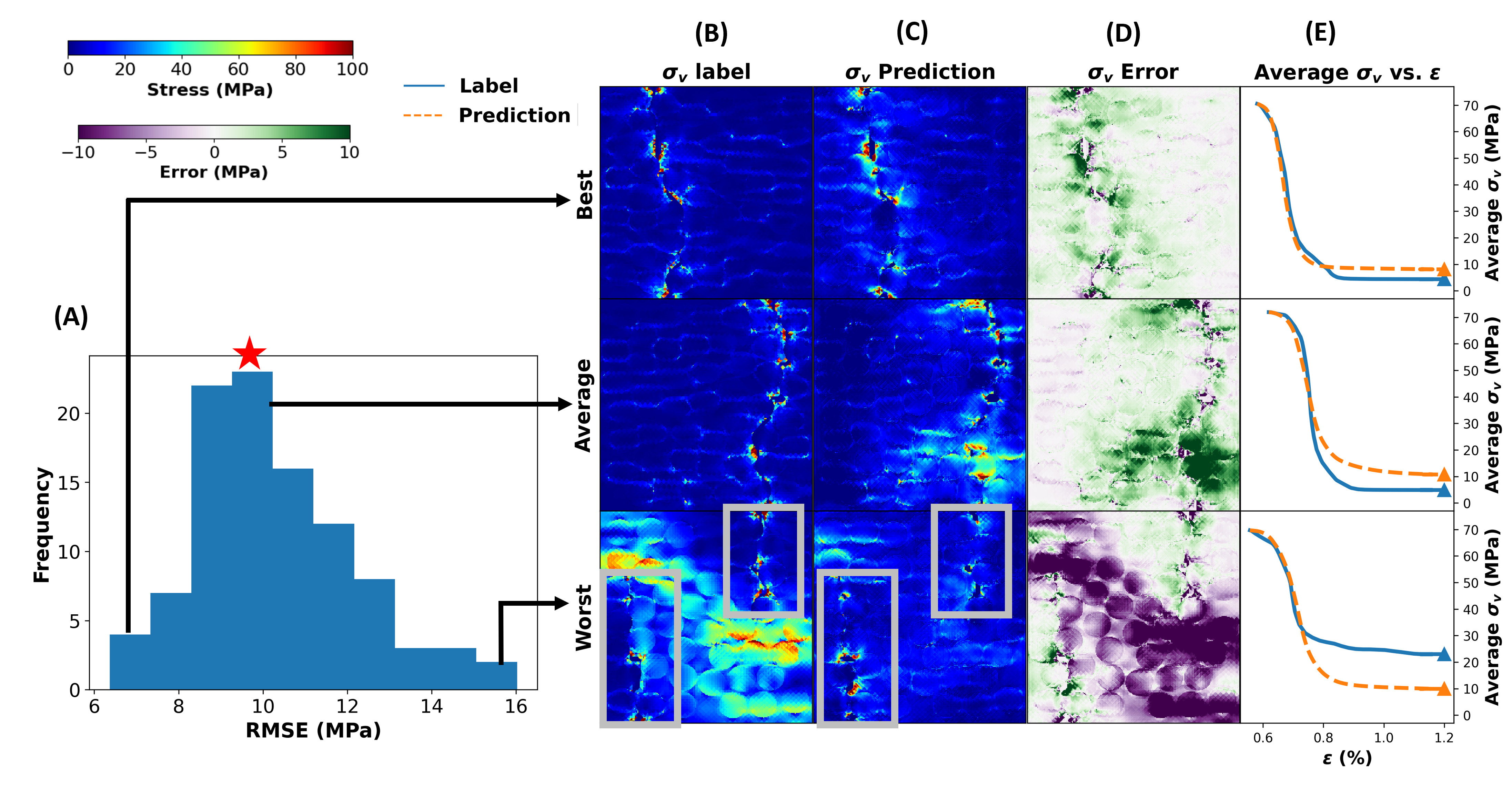}
\centering
\caption{Histogram of RMSE for the testing dataset using Necking-Net (A). The red star represents where the case in Figure~\ref{fig:Necking_t} would be in the histogram. Column B shows the ground truth stress fields at the final strain step. Column C shows the predicted stress fields at the final strain step. Column D displays the error distributions. Column E shows the average von Mises stress vs. strain curves. The triangles in Column E show where in the deformation process the von Mises stress contours are located. Each row in the columns corresponds to the best, average, and worst cases, determined by their respective RMSE values.}
\label{fig:Necking_result}
\end{figure}

Figure~\ref{fig:Necking_result} provides a statistical analysis of the von Mises stress predictions from Necking-Net, including the RMSE histogram, predicted stress fields, error maps, and average von Mises vs. strain curves for the best, average, and worst cases. The stress field predictions for the best and average cases show excellent agreement with the ground truth, with minimal deviation across the spatial domain. Even in the worst-case scenario, the predicted fields retain the core features (shown by the gray box) of the label data, demonstrating the model's robustness under varied test cases. Over 50\% of the test cases have an RMSE below 10~MPa and all test cases have an RMSE below 20~MPa, which is 10\% and 20\% of the maximum stress magnitude, respectively. The average von Mises stress vs. strain curves, shown in Column E of Figure~\ref{fig:Necking_result} illustrate the temporal prediction capabilities of Necking-Net. The best and average case predictions closely match the ground truth. The worst case under-predicted the second half of the curve but still captures the general dynamics of the necking process. These results show the ability of Necking-Net to model nonlinear spatiotemporal stress and damage post-UTS with precision.

Necking-Net is therefore well-suited for fast and accurate spatiotemporal stress and damage predictions for the post-UTS simulation. The results demonstrate predictions closely agreeing with the ground truth, highlighting its robustness across varying microstructures. Overall, Necking-Net's ability to effectively predict both spatiotemporal stress and damage makes it a valuable tool for capturing the post-UTS in Composite-Net.

\subsection{Composite-Net}
\begin{figure}[h]
\centering
\includegraphics[width=\textwidth]{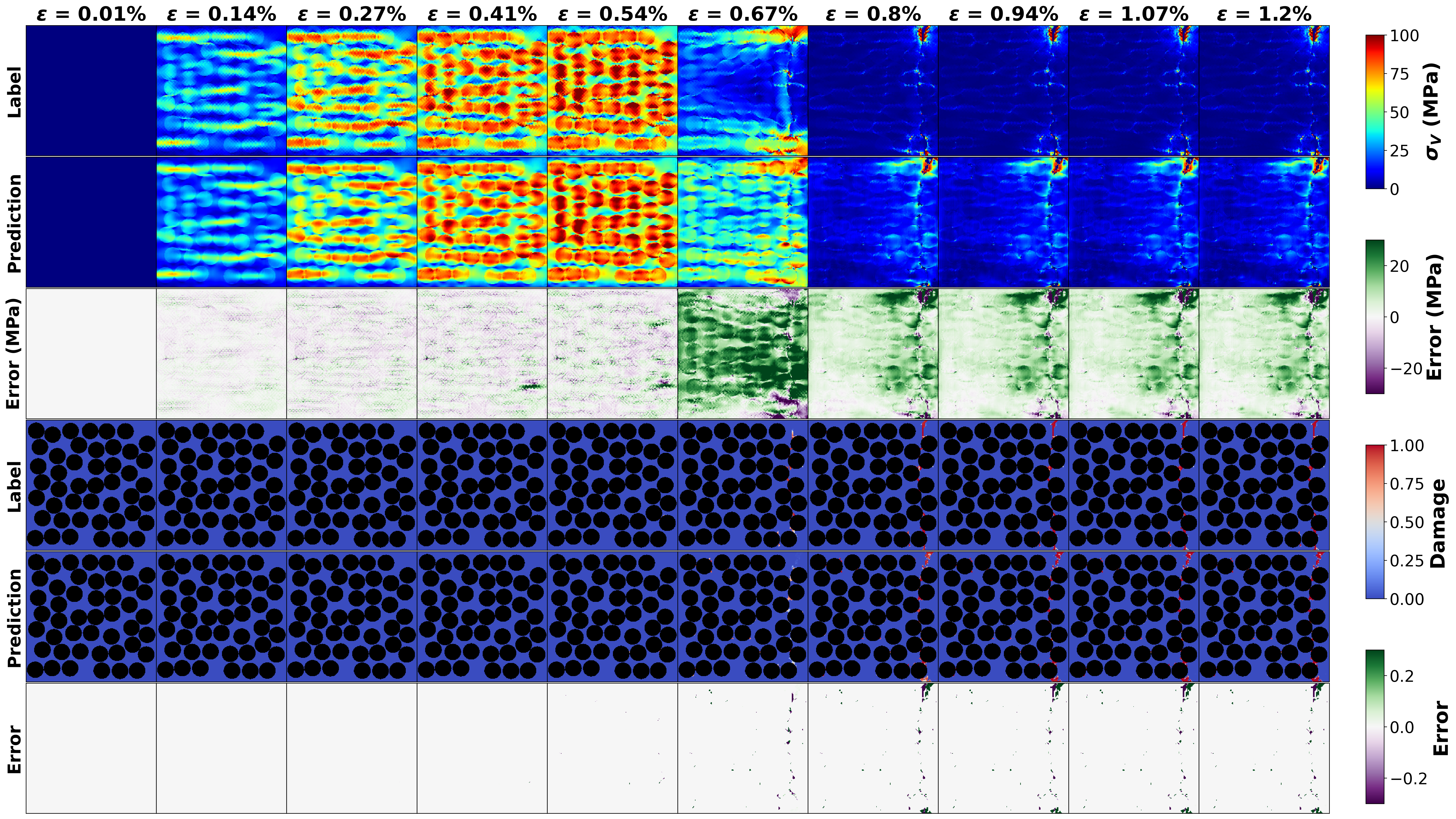}
\caption{Rollout predictions of stress (top) and damage (bottom) fields for a representative test case using the Composite-Net.}
\label{fig:composite_rollout}
\end{figure}

Composite-Net combines Damage-Net, UTS-Net, and Necking-Net into a unified predictive framework. Figure~\ref{fig:composite_rollout} illustrates the spatiotemporal ground truth and rollout predictions of von Mises stress and damage for Composite-Net of a representative test case. The predictions demonstrate the DL model's ability to seamlessly transition between the phases captured by Damage-Net, UTS-Net, and Necking-Net, maintaining spatial continuity and temporal consistency. Before UTS, the damage field remains minimal as predicted by UTS-Net, consistent with expected material behavior. As the applied strain increases, the stress and damage fields evolve incrementally, with Composite-Net accurately predicting the onset of significant damage near UTS. Post-UTS, the model captures strain localization and stress redistribution patterns, validating the performance of Necking-Net in predicting the nonlinear behavior during material softening.

\begin{figure}[h]
\centering
\includegraphics[width=\textwidth]{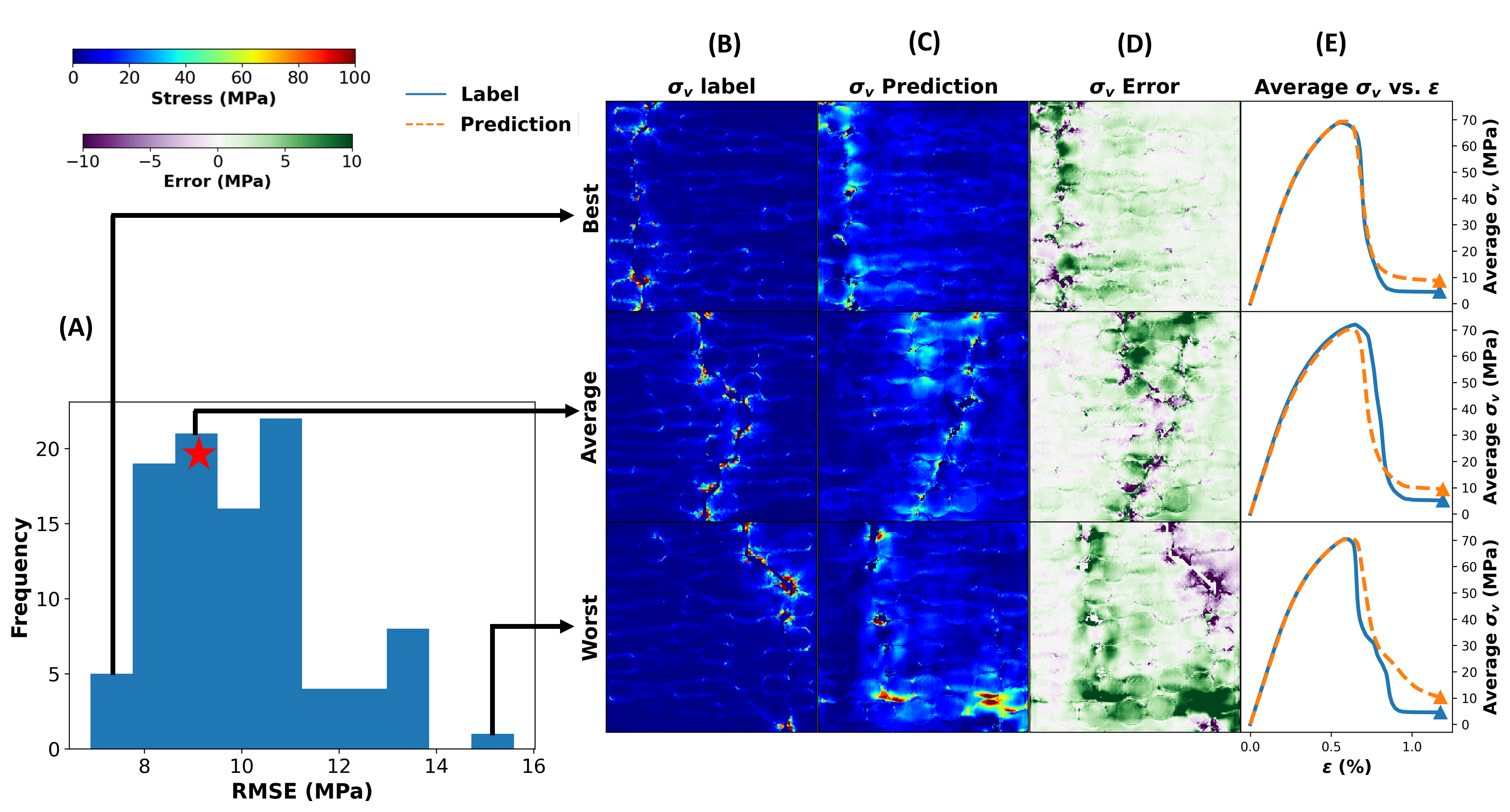}
\caption{Histogram of RMSE values for the testing dataset using the Composite-Net (A). Column B shows the ground truth stress fields at the final strain step. Column C shows the model's predicted stress field. Column D presents the error distribution, and Column E shows the average von Mises stress vs. strain curves. The triangles in Column E show where in the deformation process the von Mises stress contours are located. Each row in the columns corresponds to the best, average, and worst cases, determined by their respective RMSE values.}
\label{fig:composite_rmse}
\end{figure}

To evaluate the spatial accuracy of Composite-Net, Figure~\ref{fig:composite_rmse} presents a histogram of the RMSE for the testing dataset, along with the ground truth and predicted final von Mises stress, error distributions, and average von Mises stress vs. strain curves for the best, average, and worst cases. The best, average, and worst cases are highlighted to provide a comprehensive understanding of Composite-Net's performance. In Column B, the ground truth stress fields at the final time step are displayed. Column C shows the predicted stress fields. Column D highlights the spatial error distributions. The average von Mises stress vs. strain curves in Column E reveal the temporal accuracy of the model's predictions. 50\% of the cases have an RMSE of less than 10~MPa, and 100\% of the cases have an RMSE of less than 20~MPa, which is 10\% and 20\% of the maximum stress magnitude, respectively. For the best and average cases, the predicted stress fields closely align with the ground truth, exhibiting minimal error and capturing the spatial stress distributions effectively. The average von Mises stress vs. strain curves for these cases follow the ground truth curves, validating the model's temporal predictive capability across strain increments. Although the predictions of the worst cases show poor accuracy, they are only less than 5\% of the testing dataset, highlighting that the model performs well for the majority of cases. The best, average, and worst rollout predictions of Composite-Net are shown in ~\ref{appendix:Additional Compoiste-Net}.

The results demonstrate the capability of Composite-Net to seamlessly integrate predictions from Damage-Net, UTS-Net, and Necking-Net, enabling accurate estimation of von Mises stress and damage evolution throughout multiple stages of the loading process. Composite-Net captures the material response from the loading through damage initiation, UTS, and post-UTS deformation, providing a comprehensive representation of the progressive failure mechanisms.

In terms of computational efficiency, Composite-Net significantly outperforms IGFEM, achieving a runtime of only 8 seconds per case, compared to 490 seconds per case for IGFEM. This represents a speed-up of over 60 times, highlighting the potential of Composite-Net for rapid and scalable predictions in high-fidelity material modeling.


\section{Discussion}
The proposed composite framework, comprising Damage-Net, UTS-Net, and Necking-Net, provides a comprehensive approach for predicting the progression of stress and damage fields in composite materials under tensile loading. Each model addresses a specific phase of the material's response, from damage initiation to post-UTS behavior. This section discusses the strengths, limitations, and implications of the developed models, along with opportunities for future research.

\subsection{Model Performance and Predictive Capabilities}

The results demonstrate that Damage-Net effectively predicts the final damage field, accurately capturing critical regions where damage accumulates. By leveraging the microstructural features as inputs, the model identifies localized damage mechanisms, agreeing well with the physical behavior of composite materials under stress. UTS-Net extends these predictions by tracking the evolution of stress and damage fields up to the ultimate tensile stress. The auto-regressive framework implemented in UTS-Net allows for incremental updates to stress and damage contours, demonstrating strong temporal-spatial predictive capabilities. The average von Mises stress vs. strain curves produced by UTS-Net closely follow the ground truth, particularly in the best and average cases, confirming the model's reliability in capturing progressive material behavior. Similarly, Necking-Net successfully models the post-UTS behavior, where significant stress localization and redistribution occur. Despite the increased complexity of predicting post-UTS response, the model was able to reproduce the spatiotemporal von Mises stress and damage fields.

RMSE analysis of the unseen test cases highlights the accuracy and robustness of the predictions. By categorizing predictions into best, average, and worst cases, we identify scenarios where the models perform well and where further improvements may be needed. While the spatial stress fields exhibit high fidelity, discrepancies observed in the worst cases indicate sensitivity to certain microstructural variations.

A key advantage of Composite-Net over IGFEM is its computational efficiency. In a comparative analysis of ten cases, Using IGFEM, which utilizes CPU, required 490 seconds per case, whereas composite-net, capable of utilizing GPU, completed each case in just 8 seconds, achieving a speed-up of over 60 times. This significant reduction in computational cost highlights the practicality of Composite-Net for large-scale simulations and underscores the potential of data-driven surrogate models in accelerating high-fidelity simulations while maintaining competitive accuracy.

\subsection{Insights into Deformation and Damage Mechanisms}
One of the key contributions of this study is the ability to capture the temporal evolution of damage. Damage-Net identifies damage sites, while UTS-Net tracks the accumulation of damage up to UTS. The absence of extensive damage in the early UTS-Net predictions highlights the model's ability to capture the expected material behavior, where damage remains in the form of localized, small-scale defects that have not coalesced into a continuous crack.

Necking-Net further provides insight into the post-UTS behavior, where stress localization dominates the material response. This phase is often challenging to model due to the highly nonlinear stress-strain relationships and evolving damage patterns. The successful predictions by Necking-Net in most of the tested cases demonstrate the importance of using incremental updates to account for stress redistribution and deformation localization.

\subsection{Limitations and Future Work}
The accuracy of Composite-Net is directly dependent on the performance of Damage-Net in predicting the final damage, which contains some inaccurate cases. Improvements in surrogate models that enhance the accuracy of damage predictions will lead to a corresponding increase in the accuracy of Composite-Net.

While the proposed model demonstrating good performance in most of the cases, several limitations still remain. First, the training data for all three models are derived from FEM simulations, which inherently depend on the chosen material model and boundary conditions. The generalizability of the models to experimental data or other material systems may require additional training and validation.

Future works will focus on both extending the model's capability and accuracy. Damage-Net will be further developed because it is the bottleneck of Composite-Net. Although Damage-Net was able to predict the damage pattern correctly in most cases, there are still some outlier cases that will require model improvement. Extending the model's capability will include the framework to three-dimensional (3D) problems, where stress and damage evolution may exhibit more complex patterns. Additionally, integrating more physics-informed loss functions or constraints may further improve model accuracy by ensuring physical consistency in stress and damage predictions. Exploring transfer learning techniques to adapt the models for different loading conditions, such as compressive, cyclic and multi-axial loading, also presents a promising direction~\cite{cracknet}.

\subsection{Implications and Applications}

The ability to predict the full progression of stress and damage fields in composite materials has significant implications for material design, structural health monitoring, and failure analysis. The composite framework can serve as a surrogate model to reduce computational costs associated with traditional FEM simulations while providing rapid predictions for complex loading scenarios. This capability is particularly valuable in the design and optimization of advanced composites, where understanding the interplay between microstructure and damage evolution is critical.

Moreover, the integration of Damage-Net, UTS-Net, and Necking-Net into a unified predictive tool enables a more holistic prediction of material behavior across different deformation stages. By capturing both spatial and temporal aspects of stress and damage evolution, the proposed framework paves the way for improved predictive modeling and decision-making in engineering applications.


\section{Conclusion}
This study presents a novel U-Net-based deep learning framework for predicting the evolution of stress and damage fields within CFRC under mechanical loads. By leveraging the U-Net's capability to capture spatial and hierarchical features, the proposed model overcomes key limitations of traditional FEM-based approaches and previous machine learning models, offering a more efficient, accurate and scalable solution for predicting CFRC behavior.

The proposed framework integrates multi-scale information, demonstrating high accuracy in predicting the spatially resolving stress and damage evolution throughout the deformation process. Through its modular design—comprising UTS-Net, Damage-Net, and Necking-Net-the model effectively captures both macro-scale mechanical responses and localized microstructural phenomena, providing useful insights into the progression of material behavior and failure mechanism under loading.

The findings underscore the potential of deep learning in accelerating and enhancing the analysis and design of advanced composite materials, addressing fundamental challenges in computational efficiency. Future research will focus on expanding the framework to accommodate more complex laminate configurations, incorporating three-dimensional microstructural details, and improving generalizability across diverse loading conditions, fiber volume fractions, and spatial arrangements of fibers in CFRCs.

This work sets the stage for integrating machine learning into composite material design pipelines, enabling more reliable, efficient, and cost-effective development of advanced engineering materials for demanding applications in aerospace, automotive, and beyond.


\clearpage
\appendix

\section{High-fidelity FEM Simulation}
\label{appendix:FEM_details}
In this study, the label data used for training the deep learning model is generated using a nonlinear cohesive interface-enriched generalized finite element method. Nonlinear cohesive IGFEM is used because it enables nonconforming meshing for complex geometries with discontinuous gradient fields, such as CFRP composites. This approach simplifies meshing and reduces computational cost by allowing a uniform mesh while accurately capturing discontinuities. Additionally, nonlinear cohesive IGFEM extends traditional IGFEM by incorporating arbitrarily oriented embedded cohesive interfaces using cohesive zone models to model different constitutive behaviors within a single element~\cite{CUI2008107}. The constitutive relationships
summarized herein were previously implemented and verified by Sepasdar and Shakiba~\cite{SEPASDAR_FEM}.

In this work, the matrix is modeled using an elasto-plastic damage framework with brittle failure. An isotropic hardening rule is utilized to model the elasto-plastic response of the matrix. A strain-based rate-dependent viscous-damage model is utilized to capture the brittle failure of the matrix. The plastic deformation initiates based on the Tschoegl yield criterion as follows
\begin{equation}
\label{eq:yield_crit}
\phi = 6 J_{2} + 2 I_{1} (\sigma_c - \sigma_t) - 2 \sigma_c \sigma_t
\end{equation}
where \( J_2\) is the second invariant of the deviatoric stress tensor, which quantifies the shear stress tensor; \( I_1\) is the first invariant of the stress, representing the volumetric stress; \( \sigma_c \) is the matrix yield strength under uniaxial compression, and \( \sigma_t \) is the matrix yield strength
under uniaxial tension. 
If $\phi$ > 0, the elasto-plastic stiffness tensor \( \mathbf{D}^{ep}\) is updated using Eq.~\eqref{eq:stif_mat_calc}.

\begin{align}
\label{eq:stif_mat_calc}
\mathbf{D}^{ep} = \mathbf{D} - \frac{\left( \mathbf{D}:\frac{\partial f}{\partial \mathbf{\sigma}}\right) \otimes \left( \frac{\partial f}{\partial \mathbf{\sigma}} : \mathbf{D} \right)}{H + \frac{\partial f}{\partial\mathbf{\sigma}}:\mathbf{D}:\frac{\partial f}{\partial \mathbf{\sigma}}}
\end{align}
where \( \otimes \) is the dyadic product. \( H \) is the slope of the tangent line characterizing the transition from elastic to plastic behavior. $\mathbf{D}$ is the elastic stiffness matrix.
\( f \) is the yield function calculated as 

\begin{equation}
    f(\mathbf{\sigma}) = 6 J_{2} + 2 I_{1} (\sigma_c - \sigma_t)
\end{equation}


\( H \) is calculated using material-specific parameters \( a \) and \( b \) as 
\begin{align}
H = a \left( \frac{\sigma_Y(\epsilon_{eq}^p=0)}{\sigma_v} \right)^b
\end{align}
where $\sigma_v$ is the von Mises stress.

To assess the evolution of damage within the matrix, the Tschoegl yield criterion is calculated as follows
\begin{equation}
\label{eq:dmg_crit}
\phi' = 6 J'_{2} + 2 I'_{1} (\epsilon_c - \epsilon_t) - 2 \epsilon_c \epsilon_t
\end{equation}
where \( J_2'\) is the second invariant of the deviatoric strain, which quantifies the shear strain; \( I_1'\) is the first invariant of the strain, representing the volumetric strain; \( \epsilon_c \) and \( \epsilon_t \) are the failure strains for compression and tension, respectively. The damage parameter, initially zero, is updated incrementally according to
\begin{align}
d &= d + \frac{dt (G - Y)}{1 + \mu dt}
\label{eq:damage_update}
\end{align}
where \( dt \) represents the pseudo-time step, \( \mu \) is the fluidity coefficient, and \( Y \) is a state variable representing the damage threshold. 

The damage function, \( G \), is formulated based on the strain energy, \( \mathcal{N} \), using a damage parameter \( \overline{\tau} = \sqrt{2\mathcal{N}} \). \( G \) is expressed as 
\begin{align}
G &= 1 - \frac{\overline{\tau}_0(1 - A )}{\overline{\tau}} - A \exp\left( B (\overline{\tau}_0 - \overline{\tau}) \right),
\label{eq:damage_func}
\end{align}
where \( A \) and \( B \) are constants characterizing the material behavior, and \( \overline{\tau}_0 \) denotes the initial threshold for damage initiation. The damage threshold is updated simultaneously based on the expression
\begin{align}
Y &= \frac{Y + \mu dt G}{1 + \mu dt}
\label{eq:threshold_update}
\end{align}

Updates to \( d \) and \( Y \) are applied only when \( G > Y \). This ensures that damage evolution occurs only when the energy criterion is satisfied.

Finally, the damaged stiffness tensor, $\mathbf{D}^d$, is updated through 
\begin{align}
\mathbf{D}^d = (1 - d)\mathbf{D}^{ep} 
\end{align}

The fiber/matrix interfacial debonding was modeled using a bilinear cohesive zone model (CZM) proposed by Ortiz and Pandolfi~\cite{ortiz1999finite}. The cohesive behavior was defined based on the effective opening displacement, \( \delta = \sqrt{\delta_n^2 + \delta_t^2} \), which accounts for the interaction between normal (\(\delta_n\)) and tangential (\(\delta_t\)) debonding displacements.

The CZM parameters include fracture toughness (\( G_c \)), cohesive strength (\( T_c \)), and critical opening displacement (\( \delta_c \)). However, CZMs can lead to convergence issues in static analyses when simulating brittle debonding failures, primarily due to the Newton-Raphson iterations entering an infinite cycle~\cite{sepasdar2020overcoming}. To mitigate this issue, an artificial viscosity proposed by Gao et al.~\cite{gao2004simple} was incorporated into the cohesive law.

Table~\ref{tab:matrix_properties} and Table~\ref{tab:fiber_properties} summarizes the material properties used in the aforementioned constitutive equations for a typical CFRC.
\begin{table}[h!]
\centering
\caption{Elastic, plastic, and damage properties of the matrix material in the CFRC composite. \( E \) is Young’s modulus, \( \nu \) is Poisson’s ratio, \( \sigma_c \) and \( \sigma_t \) are compressive and tensile yield strengths, \( \epsilon_c \) and \( \epsilon_t \) are the failure strains for compression and tension, and \( A \), \( B \), \( a \), \( b \), and \( \mu \) are model-specific parameters.}
\label{tab:matrix_properties}
\begin{tabular}{|c|c|c|}
\hline
\textbf{Property} & \textbf{Value} & \textbf{Unit} \\ \hline
\( E \) & 3.9 & GPa \\ \hline
\( \nu \) & 0.39 & - \\ \hline
\( \sigma_c \) & 79 & MPa \\ \hline
\( \sigma_t \) & 62 & MPa \\ \hline
\( \epsilon_c \) & 0.35 & - \\ \hline
\( \epsilon_t \) & 0.04 & - \\ \hline
\( a \) & 20000 & - \\ \hline
\( b \) & 12 & - \\ \hline
\( A \) & 0.95 & - \\ \hline
\( B \) & 2 & - \\ \hline
\( \mu \) & 10 & - \\ \hline
\end{tabular}
\end{table}

\vspace{1em}
\begin{table}[h!]
\centering
\caption{Elastic and fracture properties of carbon fiber and fiber-matrix interactions in the CFRC composite. \( E_1 \) and \( E_2 \) are Young’s moduli in the out-of-plane and in-plane directions, \( G_{12} \) and \( G_{23} \) are shear moduli, \( \nu_{12} \) is Poisson’s ratio, \( T_c \) is the shear strength, \( \delta_c \) is the critical displacement, and \( G_c \) is the fracture energy.}
\label{tab:fiber_properties}
\begin{tabular}{|c|c|c|}
\hline
\textbf{Property} & \textbf{Value} & \textbf{Unit} \\ \hline
\( E_1 \) & 233 & GPa \\ \hline
\( E_2 \) & 23.1 & GPa \\ \hline
\( G_{12} \) & 8.96 & GPa \\ \hline
\( G_{23} \) & 8.27 & GPa \\ \hline
\( \nu_{12} \) & 0.2 & - \\ \hline
\( T_c \) & 70 & MPa \\ \hline
\( \delta_c \) & 1 & nm \\ \hline
\( G_c \) & 8.75 & N/m \\ \hline
\end{tabular}
\end{table}




\section{Additional Composite-Net spatiotemporal predictions}
\label{appendix:Additional Compoiste-Net}

\begin{figure}[h]
\centering
\includegraphics[width=\textwidth]{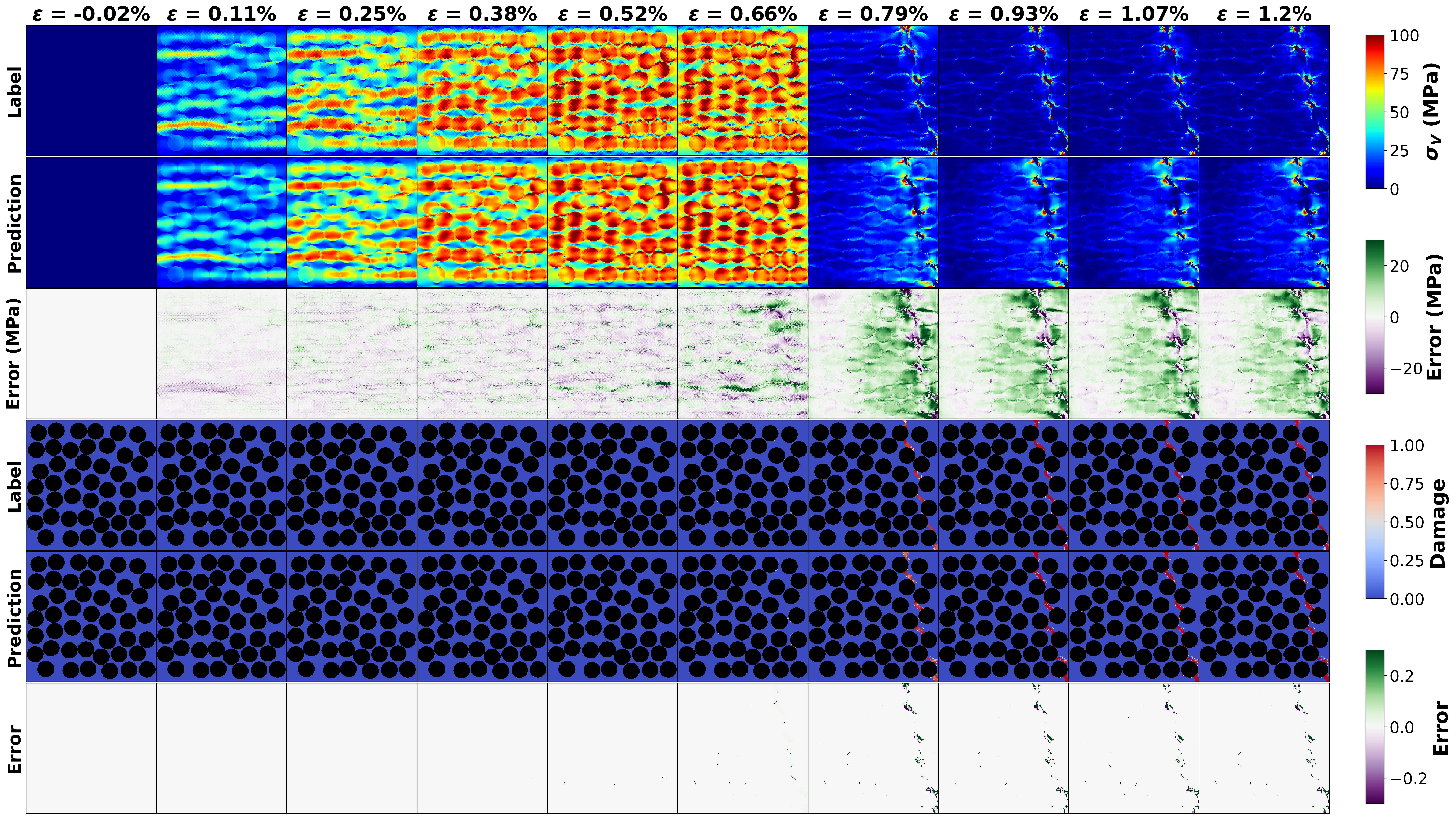}
\caption{Rollout predictions of stress (top) and damage (bottom) fields for the best test case using the Composite-Nodel.}
\label{fig:composite_rollout_best}
\end{figure}

\begin{figure}[h]
\centering
\includegraphics[width=\textwidth]{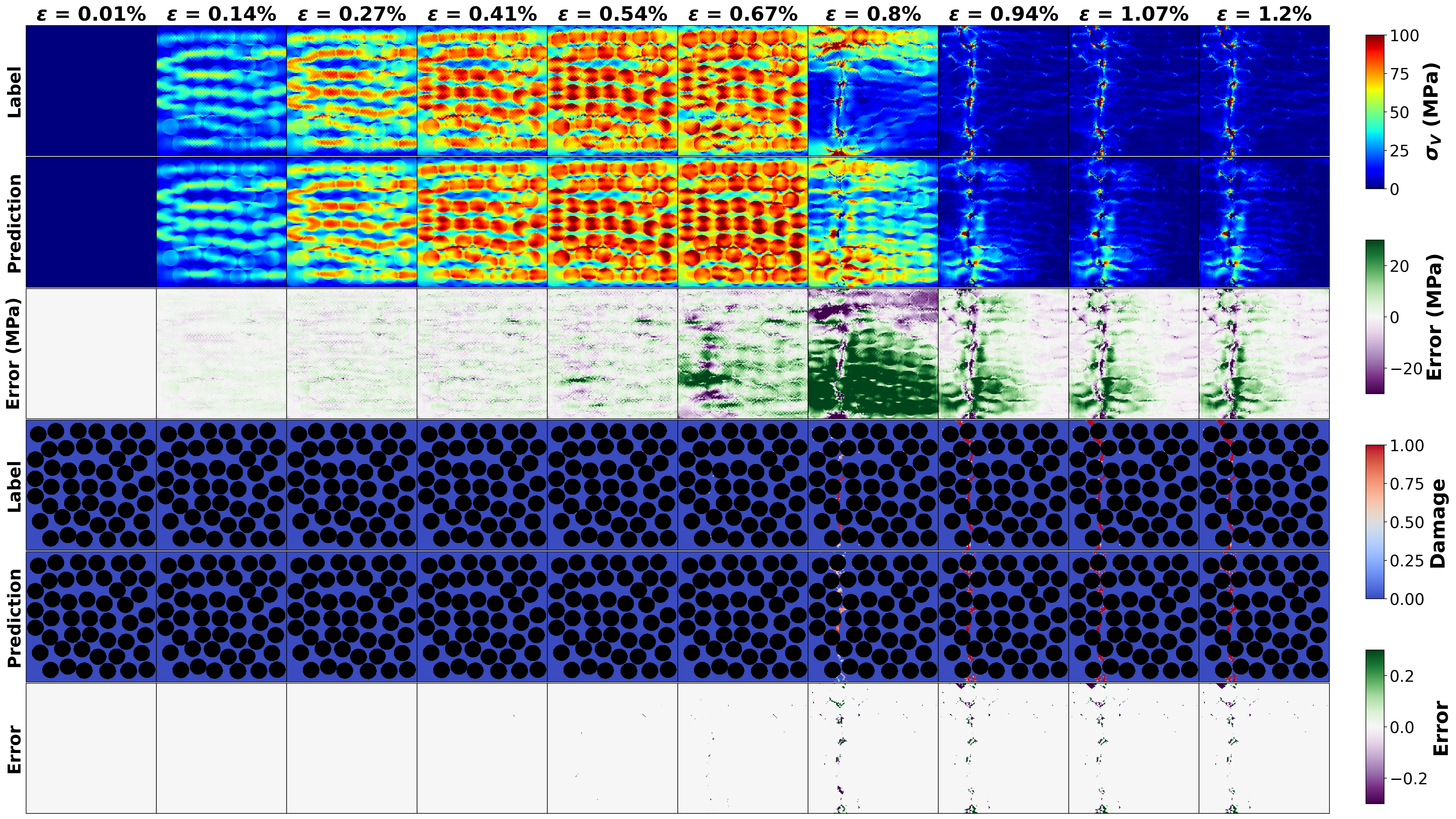}
\caption{Rollout predictions of stress (top) and damage (bottom) fields for a average test case using the Composite-Net.}
\label{fig:composite_rollout_avg}
\end{figure}

\begin{figure}[h]
\centering
\includegraphics[width=\textwidth]{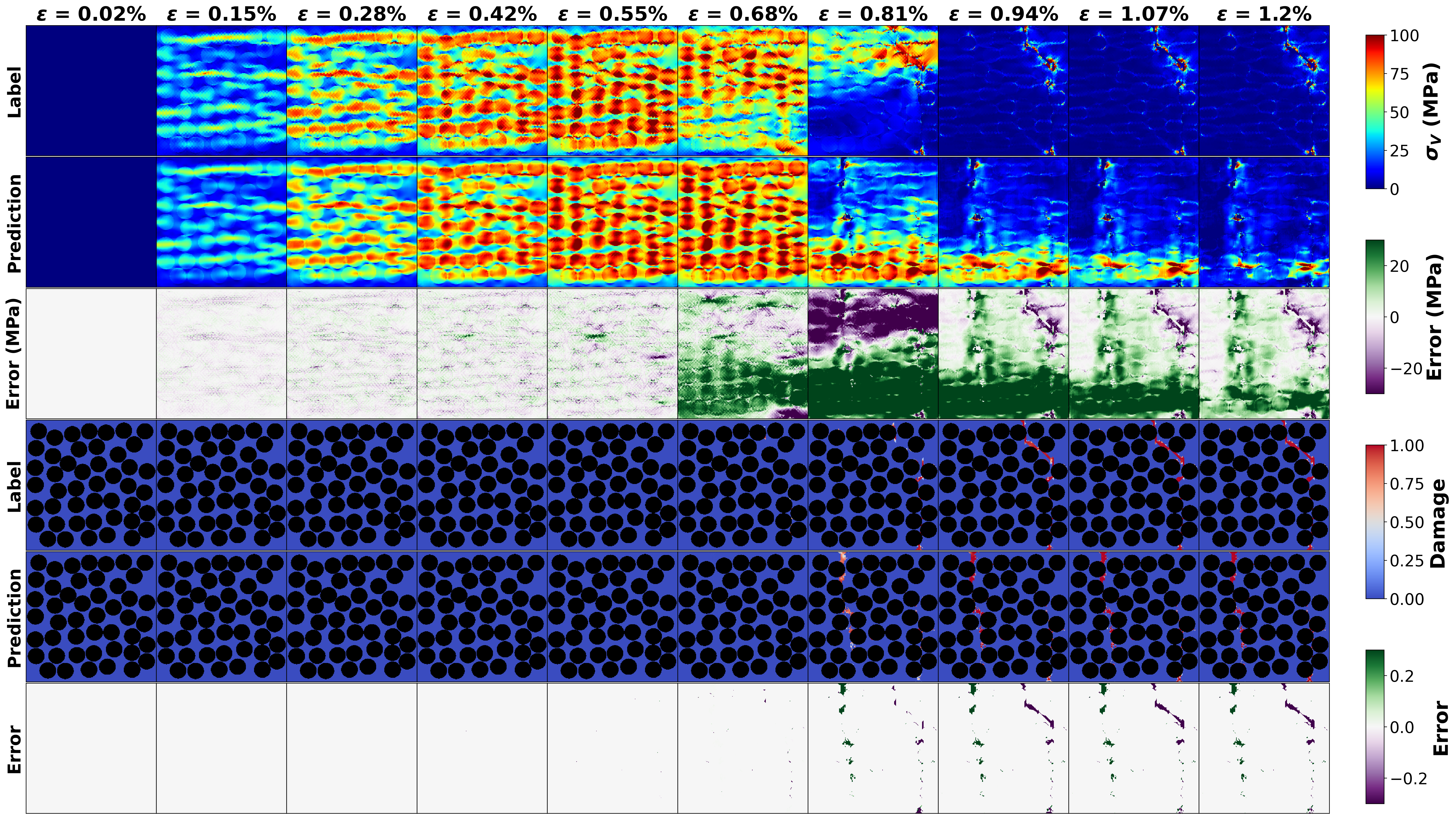}
\caption{Rollout predictions of stress (top) and damage (bottom) fields for the worst test case using the Composite-Net.}
\label{fig:composite_rollout_worst}
\end{figure}

This section shows the rollout predictions by Composite-Net. 
Figure~\ref{fig:composite_rollout_best} shows the best case in the testing dataset. Figure~\ref{fig:composite_rollout_avg} shows an average case in the testing dataset. 
Figure~\ref{fig:composite_rollout_worst} shows the worst case in the testing dataset. 

\clearpage


\section*{Data availability}
Data is provided upon request.

\section*{Code availability}
Code is provided upon request.

\section*{Acknowledgment}
The authors would like to acknowledge the funds from UES Inc. (contract no. FA8650-21-D-5279) and AFOSR (award no. FA9550-22-1-0065) in supporting this study.

\section*{Author Contributions}
Z.C., J.X.W., T.L., and V.V. all contributed to the ideation and design of the research; 
Z.C. performed the research (implemented the model, conducted numerical experiments, analyzed the data, and contributed materials/analysis tools); 
M.Y., M.S., J.X.W., T.L., and V.V. contributed to data collection and consulting. 
Z.C., M.Y., M.S., J.X.W., T.L., and V.V. contributed to manuscript editing.

\section*{Competing interests}
The authors declare no competing interests.

\noindent\textbf{Corresponding authors:} 
Vikas Varshney (\url{vikas.varshney.2@us.af.mil}), 
Jian-Xun Wang (\url{jwang33@nd.edu}), 
and Tengfei Luo (\url{tluo@nd.edu}).

\clearpage
\bibliographystyle{elsarticle-num}

\begin{thebibliography}{10}
\expandafter\ifx\csname url\endcsname\relax
  \def\url#1{\texttt{#1}}\fi
\expandafter\ifx\csname urlprefix\endcsname\relax\def\urlprefix{URL }\fi
\expandafter\ifx\csname href\endcsname\relax
  \def\href#1#2{#2} \def\path#1{#1}\fi

\bibitem{FEM}
G.~Prathap, \href{https://doi.org/10.1007/978-94-017-3319-9_1}{Introduction}, Springer Netherlands, Dordrecht, 1993, pp. 1--32.
\newblock \href {https://doi.org/10.1007/978-94-017-3319-9_1} {\path{doi:10.1007/978-94-017-3319-9_1}}.
\newline\urlprefix\url{https://doi.org/10.1007/978-94-017-3319-9_1}

\bibitem{IGFEM}
K.~Zhang, J.-M. Jin, P.~H. Geubelle, An interface-enriched generalized fem for em analysis of composites with nonconformal meshes, in: 2016 IEEE/ACES International Conference on Wireless Information Technology and Systems (ICWITS) and Applied Computational Electromagnetics (ACES), 2016, pp. 1--2.
\newblock \href {https://doi.org/10.1109/ROPACES.2016.7465468} {\path{doi:10.1109/ROPACES.2016.7465468}}.

\bibitem{MGN}
T.~Pfaff, M.~Fortunato, A.~Sanchez{-}Gonzalez, P.~W. Battaglia, \href{https://arxiv.org/abs/2010.03409}{Learning mesh-based simulation with graph networks}, CoRR abs/2010.03409 (2020).
\newblock \href {http://arxiv.org/abs/2010.03409} {\path{arXiv:2010.03409}}.
\newline\urlprefix\url{https://arxiv.org/abs/2010.03409}

\bibitem{FAN}
X.~Fan, D.~Akhare, J.-X. Wang, \href{https://www.sciencedirect.com/science/article/pii/S0045782524007321}{Neural differentiable modeling with diffusion-based super-resolution for two-dimensional spatiotemporal turbulence}, Computer Methods in Applied Mechanics and Engineering 433 (2025) 117478.
\newblock \href {https://doi.org/https://doi.org/10.1016/j.cma.2024.117478} {\path{doi:https://doi.org/10.1016/j.cma.2024.117478}}.
\newline\urlprefix\url{https://www.sciencedirect.com/science/article/pii/S0045782524007321}

\bibitem{Liu}
X.-Y. Liu, M.~H. Parikh, X.~Fan, P.~Du, Q.~Wang, Y.-F. Chen, J.-X. Wang, \href{https://arxiv.org/abs/2411.14378}{Confild-inlet: Synthetic turbulence inflow using generative latent diffusion models with neural fields} (2024).
\newblock \href {http://arxiv.org/abs/2411.14378} {\path{arXiv:2411.14378}}.
\newline\urlprefix\url{https://arxiv.org/abs/2411.14378}

\bibitem{wang2024}
Q.~Wang, P.~Ren, H.~Zhou, X.-Y. Liu, Z.~Deng, Y.~Zhang, R.~Chengze, H.~Liu, Z.~Wang, J.-X. Wang, J.-R. Wen, H.~Sun, Y.~Liu, \href{https://arxiv.org/abs/2411.00040}{P$^2$c$^2$net: Pde-preserved coarse correction network for efficient prediction of spatiotemporal dynamics} (2024).
\newblock \href {http://arxiv.org/abs/2411.00040} {\path{arXiv:2411.00040}}.
\newline\urlprefix\url{https://arxiv.org/abs/2411.00040}

\bibitem{du2024conditional}
P.~Du, M.~H. Parikh, X.~Fan, X.-Y. Liu, J.-X. Wang, Conditional neural field latent diffusion model for generating spatiotemporal turbulence, Nature Communications 15~(1) (2024) 10416.

\bibitem{Li}
R.~Li, J.~Zhou, J.-X. Wang, T.~Luo, \href{https://doi.org/10.1115/1.4067163}{Physics-informed bayesian neural networks for solving phonon boltzmann transport equation in forward and inverse problems with sparse and noisy data}, ASME Journal of Heat and Mass Transfer (2024) 1--33\href {http://arxiv.org/abs/https://asmedigitalcollection.asme.org/heattransfer/article-pdf/doi/10.1115/1.4067163/7406060/ht-24-1263.pdf} {\path{arXiv:https://asmedigitalcollection.asme.org/heattransfer/article-pdf/doi/10.1115/1.4067163/7406060/ht-24-1263.pdf}}, \href {https://doi.org/10.1115/1.4067163} {\path{doi:10.1115/1.4067163}}.
\newline\urlprefix\url{https://doi.org/10.1115/1.4067163}

\bibitem{akharediffhybrid}
D.~Akhare, T.~Luo, J.-X. Wang, \href{https://arxiv.org/abs/2401.00161}{Diffhybrid-uq: Uncertainty quantification for differentiable hybrid neural modeling} (2023).
\newblock \href {http://arxiv.org/abs/2401.00161} {\path{arXiv:2401.00161}}.
\newline\urlprefix\url{https://arxiv.org/abs/2401.00161}

\bibitem{kim2024}
S.~Kim, T.~Luo, E.~Lee, I.-S. Suh, \href{https://arxiv.org/abs/2407.20212}{Distributed quantum approximate optimization algorithm on integrated high-performance computing and quantum computing systems for large-scale optimization} (2024).
\newblock \href {http://arxiv.org/abs/2407.20212} {\path{arXiv:2407.20212}}.
\newline\urlprefix\url{https://arxiv.org/abs/2407.20212}

\bibitem{CROOM2022104191}
B.~P. Croom, M.~Berkson, R.~K. Mueller, M.~Presley, S.~Storck, \href{https://www.sciencedirect.com/science/article/pii/S0167663621004026}{Deep learning prediction of stress fields in additively manufactured metals with intricate defect networks}, Mechanics of Materials 165 (2022) 104191.
\newblock \href {https://doi.org/https://doi.org/10.1016/j.mechmat.2021.104191} {\path{doi:https://doi.org/10.1016/j.mechmat.2021.104191}}.
\newline\urlprefix\url{https://www.sciencedirect.com/science/article/pii/S0167663621004026}

\bibitem{CVI}
D.~Akhare, Z.~Chen, R.~Gulotty, T.~Luo, J.-X. Wang, Probabilistic physics-integrated neural differentiable modeling for isothermal chemical vapor infiltration process, npj Computational Materials 10~(1) (2024) 120.

\bibitem{PINDIFF}
D.~Akhare, T.~Luo, J.-X. Wang, \href{https://www.sciencedirect.com/science/article/pii/S0045782523000257}{Physics-integrated neural differentiable (pindiff) model for composites manufacturing}, Computer Methods in Applied Mechanics and Engineering 406 (2023) 115902.
\newblock \href {https://doi.org/https://doi.org/10.1016/j.cma.2023.115902} {\path{doi:https://doi.org/10.1016/j.cma.2023.115902}}.
\newline\urlprefix\url{https://www.sciencedirect.com/science/article/pii/S0045782523000257}

\bibitem{Luo_additive}
S.~M. Estalaki, C.~S. Lough, R.~G. Landers, E.~C. Kinzel, T.~Luo, \href{https://www.sciencedirect.com/science/article/pii/S2214860422004018}{Predicting defects in laser powder bed fusion using in-situ thermal imaging data and machine learning}, Additive Manufacturing 58 (2022) 103008.
\newblock \href {https://doi.org/https://doi.org/10.1016/j.addma.2022.103008} {\path{doi:https://doi.org/10.1016/j.addma.2022.103008}}.
\newline\urlprefix\url{https://www.sciencedirect.com/science/article/pii/S2214860422004018}

\bibitem{Michopoulos}
J.~G. Michopoulos, A.~Bhaduri, F.~Chinesta, E.~Cueto, D.~Liu, S.~K. Ravi, J.-X. Wang, \href{https://doi.org/10.1115/1.4066791}{Special issue: Scientific machine learning for manufacturing processes and material systems}, Journal of Computing and Information Science in Engineering 24~(11) (2024) 110301.
\newblock \href {http://arxiv.org/abs/https://asmedigitalcollection.asme.org/computingengineering/article-pdf/24/11/110301/7387063/jcise\_24\_11\_110301.pdf} {\path{arXiv:https://asmedigitalcollection.asme.org/computingengineering/article-pdf/24/11/110301/7387063/jcise\_24\_11\_110301.pdf}}, \href {https://doi.org/10.1115/1.4066791} {\path{doi:10.1115/1.4066791}}.
\newline\urlprefix\url{https://doi.org/10.1115/1.4066791}

\bibitem{Maurizi}
M.~Maurizi, C.~Gao, F.~Berto, Predicting stress, strain and deformation fields in materials and structures with graph neural networks, Scientific Reports 12 (12 2022).
\newblock \href {https://doi.org/10.1038/s41598-022-26424-3} {\path{doi:10.1038/s41598-022-26424-3}}.

\bibitem{yacouti_integrated_2025}
M.~Yacouti, M.~Shakiba, \href{https://www.sciencedirect.com/science/article/pii/S1359835X2400616X}{Integrated convolutional and graph neural networks for predicting mechanical fields in composite microstructures}, Composites Part A: Applied Science and Manufacturing 190 (2025) 108618.
\newblock \href {https://doi.org/10.1016/j.compositesa.2024.108618} {\path{doi:10.1016/j.compositesa.2024.108618}}.
\newline\urlprefix\url{https://www.sciencedirect.com/science/article/pii/S1359835X2400616X}

\bibitem{khorrami}
M.~S. Khorrami, J.~R. Mianroodi, N.~H. Siboni, P.~Goyal, B.~Svendsen, P.~Benner, D.~Raabe, \href{https://arxiv.org/abs/2208.13490}{An artificial neural network for surrogate modeling of stress fields in viscoplastic polycrystalline materials} (2022).
\newblock \href {http://arxiv.org/abs/2208.13490} {\path{arXiv:2208.13490}}.
\newline\urlprefix\url{https://arxiv.org/abs/2208.13490}

\bibitem{Yacouti}
M.~Yacouti, M.~Shakiba, \href{https://doi.org/10.1007/s00366-024-01966-4}{Performance evaluation of deep learning approaches for predicting mechanical fields in composites}, Eng. with Comput. 40~(5) (2024) 3073–3086.
\newblock \href {https://doi.org/10.1007/s00366-024-01966-4} {\path{doi:10.1007/s00366-024-01966-4}}.
\newline\urlprefix\url{https://doi.org/10.1007/s00366-024-01966-4}

\bibitem{SEPASDAR_deep_learning}
R.~Sepasdar, A.~Karpatne, M.~Shakiba, \href{https://www.sciencedirect.com/science/article/pii/S0045782522003085}{A data-driven approach to full-field nonlinear stress distribution and failure pattern prediction in composites using deep learning}, Computer Methods in Applied Mechanics and Engineering 397 (2022) 115126.
\newblock \href {https://doi.org/https://doi.org/10.1016/j.cma.2022.115126} {\path{doi:https://doi.org/10.1016/j.cma.2022.115126}}.
\newline\urlprefix\url{https://www.sciencedirect.com/science/article/pii/S0045782522003085}

\bibitem{YAN2024110278}
H.~Yan, H.~Yu, S.~Zhu, Y.~Yin, L.~Guo, \href{https://www.sciencedirect.com/science/article/pii/S0013794424004417}{Machine learning based framework for rapid forecasting of the crack propagation}, Engineering Fracture Mechanics 307 (2024) 110278.
\newblock \href {https://doi.org/https://doi.org/10.1016/j.engfracmech.2024.110278} {\path{doi:https://doi.org/10.1016/j.engfracmech.2024.110278}}.
\newline\urlprefix\url{https://www.sciencedirect.com/science/article/pii/S0013794424004417}

\bibitem{xu}
H.~Xu, W.~Fan, A.~C. Taylor, D.~Zhang, L.~Ruan, R.~Shi, \href{https://arxiv.org/abs/2309.13626}{Crack-net: Prediction of crack propagation in composites} (2023).
\newblock \href {http://arxiv.org/abs/2309.13626} {\path{arXiv:2309.13626}}.
\newline\urlprefix\url{https://arxiv.org/abs/2309.13626}

\bibitem{UNET}
O.~Ronneberger, P.~Fischer, T.~Brox, \href{http://arxiv.org/abs/1505.04597}{U-net: Convolutional networks for biomedical image segmentation}, CoRR abs/1505.04597 (2015).
\newblock \href {http://arxiv.org/abs/1505.04597} {\path{arXiv:1505.04597}}.
\newline\urlprefix\url{http://arxiv.org/abs/1505.04597}

\bibitem{SEPASDAR_FEM}
R.~Sepasdar, M.~Shakiba, \href{https://www.sciencedirect.com/science/article/pii/S0263822321014112}{Micromechanical study of multiple transverse cracking in cross-ply fiber-reinforced composite laminates}, Composite Structures 281 (2022) 114986.
\newblock \href {https://doi.org/https://doi.org/10.1016/j.compstruct.2021.114986} {\path{doi:https://doi.org/10.1016/j.compstruct.2021.114986}}.
\newline\urlprefix\url{https://www.sciencedirect.com/science/article/pii/S0263822321014112}

\bibitem{pyvista}
Q.~Liu, Z.~Qiao, Y.~Lv, \href{https://www.sciencedirect.com/science/article/pii/S1270963821004715}{Pyvt: A python-based open-source software for visualization and graphic analysis of fluid dynamics datasets}, Aerospace Science and Technology 117 (2021) 106961.
\newblock \href {https://doi.org/https://doi.org/10.1016/j.ast.2021.106961} {\path{doi:https://doi.org/10.1016/j.ast.2021.106961}}.
\newline\urlprefix\url{https://www.sciencedirect.com/science/article/pii/S1270963821004715}

\bibitem{Ensemble}
M.~Ganaie, M.~Hu, A.~Malik, M.~Tanveer, P.~Suganthan, \href{http://dx.doi.org/10.1016/j.engappai.2022.105151}{Ensemble deep learning: A review}, Engineering Applications of Artificial Intelligence 115 (2022) 105151.
\newblock \href {https://doi.org/10.1016/j.engappai.2022.105151} {\path{doi:10.1016/j.engappai.2022.105151}}.
\newline\urlprefix\url{http://dx.doi.org/10.1016/j.engappai.2022.105151}

\bibitem{cracknet}
H.~Xu, W.~Fan, A.~C. Taylor, D.~Zhang, L.~Ruan, R.~Shi, \href{https://arxiv.org/abs/2309.13626}{Crack-net: Prediction of crack propagation in composites} (2023).
\newblock \href {http://arxiv.org/abs/2309.13626} {\path{arXiv:2309.13626}}.
\newline\urlprefix\url{https://arxiv.org/abs/2309.13626}

\bibitem{CUI2008107}
W.~C. CUI, \href{https://www.sciencedirect.com/science/article/pii/B978184569334350005X}{5 - fatigue cracking in aged structures}, in: J.~Paik, R.~Melchers (Eds.), Condition Assessment of Aged Structures, Woodhead Publishing Series in Civil and Structural Engineering, Woodhead Publishing, 2008, pp. 107--148.
\newblock \href {https://doi.org/https://doi.org/10.1533/9781845695217.2.107} {\path{doi:https://doi.org/10.1533/9781845695217.2.107}}.
\newline\urlprefix\url{https://www.sciencedirect.com/science/article/pii/B978184569334350005X}

\bibitem{ortiz1999finite}
M.~Ortiz, A.~Pandolfi, Finite-deformation irreversible cohesive elements for three-dimensional crack-propagation analysis, International Journal for Numerical Methods in Engineering 44~(9) (1999) 1267--1282.

\bibitem{sepasdar2020overcoming}
R.~Sepasdar, M.~Shakiba, Overcoming the convergence difficulty of cohesive zone models through a {Newton-Raphson} modification technique, Engineering Fracture Mechanics (2020) 107046.

\bibitem{gao2004simple}
Y.~Gao, A.~Bower, A simple technique for avoiding convergence problems in finite element simulations of crack nucleation and growth on cohesive interfaces, Modelling and Simulation in Materials Science and Engineering 12~(3) (2004) 453.

\end{thebibliography}

\end{document}